\pdfoutput=1

\documentclass[11pt]{article}

\usepackage[]{EMNLP}

\usepackage{times}
\usepackage{latexsym}

\usepackage[T1]{fontenc}
\usepackage[utf8]{inputenc}
\usepackage{microtype}
\usepackage{inconsolata}

\usepackage{tikz}
\usetikzlibrary{positioning, fit, calc}

\usepackage{amsmath}
\usepackage{dblfloatfix}
\usepackage{booktabs, multirow}

\usepackage{graphicx}
\usepackage{subcaption}
\usepackage{dblfloatfix} 

\usepackage{algorithm}
\usepackage{algorithmic}

\usepackage{tikz}
\usetikzlibrary{decorations.pathreplacing}

\title{Linguistically Motivated Sign Language Segmentation}

\author{
\parbox{0.6\linewidth}{\centering
Amit Moryossef$^{*1,2}$, Zifan Jiang$^{*2}$ \\
Mathias Müller$^2$, Sarah Ebling$^2$, Yoav Goldberg$^1$}
\\
$^1$Bar-Ilan University, 
$^2$University of Zurich
\\
\texttt{amitmoryossef@gmail.com}, \texttt{jiang@cl.uzh.ch}}

\begin{document}

\maketitle
\def\thefootnote{*}\footnotetext{Equal contribution authors.}
\def\thefootnote{\arabic{footnote}}

\begin{abstract}

Sign language segmentation is a crucial task in sign language processing systems. It enables downstream tasks such as sign recognition, transcription, and machine translation.
In this work, we consider two kinds of segmentation: segmentation into individual signs and segmentation into \textit{phrases}, larger units comprising several signs. We propose a novel approach to jointly model these two tasks.

Our method is motivated by linguistic cues observed in sign language corpora. We replace the predominant IO tagging scheme with BIO tagging to account for continuous signing. Given that prosody plays a significant role in phrase boundaries, we explore the use of optical flow features. We also provide an extensive analysis of hand shapes and 3D hand normalization.

We find that introducing BIO tagging is necessary to model sign boundaries. 
Explicitly encoding prosody by optical flow improves segmentation in shallow models, but its contribution is negligible in deeper models.
Careful tuning of the decoding algorithm atop the models further improves the segmentation quality.

We demonstrate that our final models generalize to out-of-domain video content in a different signed language, even under a zero-shot setting.
We observe that including optical flow and 3D hand normalization enhances the robustness of the model in this context.

\end{abstract}

\section{Introduction}\label{sec:introduction}

Signed languages are natural languages used by deaf and hard-of-hearing individuals to communicate through a combination of manual and non-manual elements \citep{sandler2006sign}. 
Like spoken languages, signed languages have their own distinctive grammar, and vocabulary, that have evolved through natural processes of language development \citep{sandler2010prosody}. 

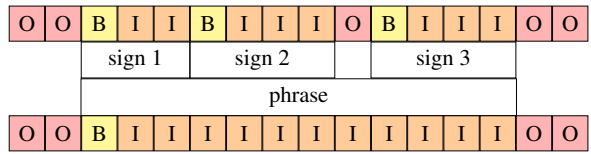
\begin{figure}[ht]

\resizebox{\linewidth}{!}{%
\begin{tikzpicture}[Obox/.style={draw, rectangle, minimum size=0.6cm, fill=red!30}, 
Bbox/.style={draw, rectangle, minimum size=0.6cm, fill=yellow!50}, 
Ibox/.style={draw, rectangle, minimum size=0.6cm, fill=orange!40}, 
sign/.style={draw, rectangle, inner sep=0pt},
node distance=0cm]

% First line of letters in boxes
\node[Obox] (o1) {O};
\node[Obox,right=of o1] (o2) {O};
\node[Bbox,right=of o2] (b1) {B};
\node[Ibox,right=of b1] (i1) {I};
\node[Ibox,right=of i1] (i2) {I};
\node[Bbox,right=of i2] (b2) {B};
\node[Ibox,right=of b2] (i3) {I};
\node[Ibox,right=of i3] (i4) {I};
\node[Ibox,right=of i4] (i5) {I};
\node[Obox,right=of i5] (o4) {O};
\node[Bbox,right=of o4] (b3) {B};
\node[Ibox,right=of b3] (i6) {I};
\node[Ibox,right=of i6] (i7) {I};
\node[Ibox,right=of i7] (i8) {I};
\node[Obox,right=of i8] (o5) {O};
\node[Obox,right=of o5] (o6) {O};

% Box with three signs
\node[sign, fit=(b1) (i2), below=0cm of $(b1.south)!0.5!(i2.south)$, label=center:{sign 1}] (s1) {};
\node[sign, fit=(b2) (i5), below=0cm of $(b2.south)!0.5!(i5.south)$, label=center:{sign 2}] (s2) {};
\node[sign, fit=(b3) (i8), below=0cm of $(b3.south)!0.5!(i8.south)$, label=center:{sign 3}] (s3) {};

% Phrase box
\node[sign, fit=(s1) (s3), below=0cm of $(s1.south)!0.53!(s3.south)$, label=center:{phrase}] (phrase) {};

% Second line of letters in boxes
\node[Obox,below=1.26cm of o1] (o8) {O};
\node[Obox,right=of o8] (o9) {O};
\node[Bbox,right=of o9] (b4) {B};
\node[Ibox,right=of b4] (i9) {I};
\node[Ibox,right=of i9] (i10) {I};
\node[Ibox,right=of i10] (i11) {I};
\node[Ibox,right=of i11] (i12) {I};
\node[Ibox,right=of i12] (i13) {I};
\node[Ibox,right=of i13] (i14) {I};
\node[Ibox,right=of i14] (i15) {I};
\node[Ibox,right=of i15] (i16) {I};
\node[Ibox,right=of i16] (i17) {I};
\node[Ibox,right=of i17] (i18) {I};
\node[Ibox,right=of i18] (i19) {I};
\node[Obox,right=of i19] (o11) {O};
\node[Obox,right=of o11] (o12) {O};
\end{tikzpicture}
}

\caption{Per-frame classification of a sign language utterance following a BIO tagging scheme. Each box represents a single frame of a video. We propose a joint model to segment \emph{signs} (top) and \emph{phrases} (bottom) at the same time. B=beginning, I=inside, O=outside. The figure illustrates continuous signing where signs often follow each other without an O frame between them.}
\label{fig:bio-example}

\end{figure}
\begin{figure*}[!t]

\begin{subfigure}{\textwidth}
\includegraphics[width=\textwidth]{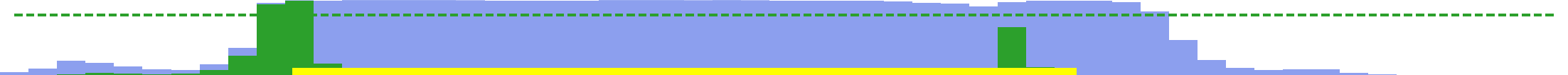}%
\vspace{-1pt}
\end{subfigure}
\begin{subfigure}{\textwidth}
\foreach \i in {11,16,...,66} {\includegraphics[width=0.083\textwidth]{figures/segments/video/00\i.png}}%
\vspace{-1pt}
\end{subfigure}
\begin{subfigure}{\textwidth}
\reflectbox{\includegraphics[width=\textwidth,angle=180,origin=c]{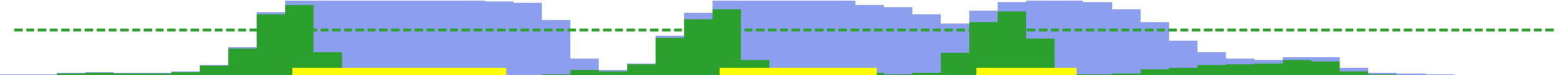}}
\end{subfigure}

\caption{The annotation of the first phrase in a video from the test set (\href{https://www.sign-lang.uni-hamburg.de/meinedgs/html/1247641_en.html}{dgskorpus\_goe\_02}), in yellow, signing: ``Why do you smoke?'' through the use of three signs: \emph{WHY} (+mouthed), \emph{TO-SMOKE}, and a gesture (+mouthed) towards the other signer.
At the top, our phrase segmentation model predicts a single phrase that initiates with a B tag (in green) above the B-threshold (green dashed line), followed by an I (in light blue), and continues until falling below a certain threshold.
At the bottom, our sign segmentation model accurately segments the three signs.}
\label{fig:main-fig-dgs}

\end{figure*}

Sign language transcription and translation systems rely on the accurate temporal segmentation of sign language videos into meaningful units such as signs \citep{segmentation:santemiz2009automatic, segmentation:renz2021signa} or signing sequences corresponding to subtitle units\footnote{Subtitles may not always correspond directly to sentences. They frequently split within a sentence and could be temporally offset from the corresponding signing segments.}
\citep{segmentation:bull2020automatic}.
However, sign language segmentation remains a challenging task due to the difficulties in defining meaningful units in signed languages \citep{segmentation:de-sisto-etal-2021-defining}.
Our approach is the first to consider two kinds of units in one model. We simultaneously segment single signs and phrases (larger units) in a unified framework.

Previous work typically approached segmentation as a binary classification task (including segmentation tasks in audio signal processing and computer vision), where each frame/pixel is predicted to be either part of a segment or not. However, this approach neglects the intricate nuances of continuous signing, where segment boundaries are not strictly binary and often blend in reality. One sign or phrase can immediately follow another, transitioning smoothly, without a frame between them being distinctly \emph{outside} (Figure \ref{fig:bio-example} and \S\ref{sec:motivation-bio}). 

We propose incorporating linguistically motivated cues to address these challenges and improve sign language segmentation.
To cope with continuous signing, we adopt a BIO-tagging approach \citep{ramshaw-marcus-1995-text}, where in addition to predicting a frame to be \emph{in} or \emph{out} of a segment, we also classify the \emph{beginning} of the segment as shown in Figure \ref{fig:main-fig-dgs}.
Since phrase segmentation is primarily marked with prosodic cues (i.e., pauses, extended sign duration, facial expressions) \citep{sandler2010prosody, ormel2012prosodic}, we explore using optical flow to explicitly model them (\S\ref{sec:motivation-optical-flow}). 
Since signs employ a limited number of hand shapes, we additionally perform 3D hand normalization (\S\ref{sec:motivation-hand-shapes}). % to facilitate sign boundary detection.

Evaluating on the Public DGS Corpus \citep{dataset:prillwitz2008dgs, dataset:hanke-etal-2020-extending} (DGS stands for German Sign Language), we report enhancements in model performance following specific modifications. 
% Our modifications 
% Implementing bidirectionality in our recurrent encoder induced considerable improvements, demonstrating the valuable contributions of context from both directions for the segmentation of sign language. 
%The adoption of BIO tagging further refined our results, allowing for the distinction between the beginning and inside of signed expressions, and therefore better boundary estimation.
%Interestingly, the explicit incorporation of pauses, modeled by optical flow, only improves segmentation performance in shallow models.
We compare our final models after hyperparameter optimization, including parameters for the decoding algorithm, and find that our best architecture using only the poses is comparable to the one that uses optical flow and hand normalization.

Reassuringly, we find that our model generalizes when evaluated on additional data from different signed languages in a zero-shot approach. We obtain segmentation scores that are competitive with previous work and observe that incorporating optical flow and hand normalization makes the model more robust for out-of-domain data.

Lastly, we conduct an extensive analysis of pose-based hand manipulations for signed languages (Appendix \ref{appendix:hand-analysis}). 
Despite not improving our segmentation model due to noise from current 3D pose estimation models, we emphasize its potential value for future work involving skeletal hand poses. Based on this analysis, we propose several measurable directions for improving 3D pose estimation.

% In the remainder of this work, we review related work on sign language segmentation in \S\ref{sec:related-work}, explain our linguistic motivation in \S\ref{sec:motivation}, describe our proposed methodology in \S\ref{sec:setup}, present our results and evaluate the performance of our approach in \S\ref{sec:results}, and conclude the work in \S\ref{sec:conclusion}.

% Reproduction code and models are available at \url{https://github.com/anon/anonymous}.

Our code and models are available at \url{https://github.com/sign-language-processing/transcription}.

\section{Related Work}\label{sec:related-work}

\subsection{Sign Language Detection}
% \subsection{Sign Language Detection}\label{tasks-sign-language-detection}

Sign language detection
\citep{detection:borg2019sign, detection:moryossef2020real, detection:pal2023importance}
is the task of determining whether signing activity is present in a given video frame.
A similar task in spoken languages is voice activity detection (VAD)
\citep{sohn1999statistical, ramirez2004efficient}, the detection of when human voice is used in an audio signal. As VAD methods often
rely on speech-specific representations such as spectrograms, they are not necessarily applicable to videos.

\citet{detection:borg2019sign} introduced the classification of frames
taken from YouTube videos as either signing or not signing. They took a
spatial and temporal approach based on VGG-16 \citep{simonyan2015very}
CNN to encode each frame and used a Gated Recurrent Unit (GRU)
\citep{cho2014learning} to encode the sequence of frames in a window of
20 frames at 5fps. In addition to the raw frame, they either encoded
optical-flow history, aggregated motion history, or frame difference.

\citet{detection:moryossef2020real} improved upon their method by
performing sign language detection in real time. They identified that
sign language use involves movement of the body and, as such, designed a
model that works on top of estimated human poses rather than directly on
the video signal. They calculated the optical flow norm of every joint
detected on the body and applied a shallow yet effective contextualized
model to predict for every frame whether the person is signing or not.

While these recent detection models achieve high performance, we need well-annotated data including interference and non-signing distractions for proper real-world evaluation.
\citet{detection:pal2023importance} conducted a detailed analysis of the
impact of signer overlap between the training and test sets on two sign
detection benchmark datasets (Signing in the Wild
\citep{detection:borg2019sign} and the DGS Corpus
\citep{dataset:hanke-etal-2020-extending}) used by
\citet{detection:borg2019sign} and \citet{detection:moryossef2020real}.
By comparing the accuracy with and without overlap, they noticed a
relative decrease in performance for signers not present during
training. As a result, they suggested new dataset partitions that
eliminate overlap between train and test sets and facilitate a more
accurate evaluation of performance.

\subsection{Sign Language Segmentation}
% \subsection{Sign Language Segmentation}\label{tasks-sign-language-segmentation}

Segmentation consists of detecting the frame boundaries for signs or
phrases in videos to divide them into meaningful units. While the most
canonical way of dividing a spoken language text is into a linear
sequence of words, due to the simultaneity of sign language, the notion
of a sign language ``word'' is ill-defined, and sign language cannot be
fully linearly modeled.

Current methods resort to segmenting units loosely mapped to signed
language units
\citep{segmentation:santemiz2009automatic, segmentation:farag2019learning, segmentation:bull2020automatic, segmentation:renz2021signa, segmentation:renz2021signb, segmentation:bull2021aligning}
and do not explicitly leverage reliable linguistic predictors of sentence
boundaries such as prosody in signed languages (i.e., pauses, extended sign duration, facial expressions)
\citep{sandler2010prosody, ormel2012prosodic}.
\citet{segmentation:de-sisto-etal-2021-defining} call for a better
understanding of sign language structure, which they believe is the
necessary ground for the design and development of sign language
recognition and segmentation methodologies.

\citet{segmentation:santemiz2009automatic} automatically extracted
isolated signs from continuous signing by aligning the sequences
obtained via speech recognition, modeled by Dynamic Time Warping (DTW)
and Hidden Markov Models (HMMs) approaches.

\citet{segmentation:farag2019learning} used a random forest classifier
to distinguish frames containing signs in Japanese Sign Language based
on the composition of spatio-temporal angular and distance features
between domain-specific pairs of joint segments.

\citet{segmentation:bull2020automatic} segmented French Sign Language
into segments corresponding to subtitle units by relying on the alignment between subtitles
and sign language videos, leveraging a spatio-temporal graph
convolutional network (STGCN; \citet{Yu2017SpatioTemporalGC}) with a BiLSTM on 2D skeleton data.

\citet{segmentation:renz2021signa} located temporal
boundaries between signs in continuous sign language videos by employing
3D convolutional neural network representations with iterative temporal
segment refinement to resolve ambiguities between sign boundary cues.
\citet{segmentation:renz2021signb} further proposed the
Changepoint-Modulated Pseudo-Labelling (CMPL) algorithm to solve the
problem of source-free domain adaptation.

\citet{segmentation:bull2021aligning} presented a Transformer-based
approach to segment sign language videos and align them with subtitles
simultaneously, encoding subtitles by BERT \citep{devlin-etal-2019-bert} and videos by CNN video
representations.

\section{Motivating Observations}\label{sec:motivation}

To motivate our proposed approach, we make a series of observations regarding the intrinsic nature of sign language expressions. Specifically, we highlight the unique challenges posed by the continuous flow of sign language expressions (\S\ref{sec:motivation-bio}), the role of prosody in determining phrase boundaries (\S\ref{sec:motivation-optical-flow}), and the influence of hand shape changes in indicating sign boundaries (\S\ref{sec:motivation-hand-shapes}).

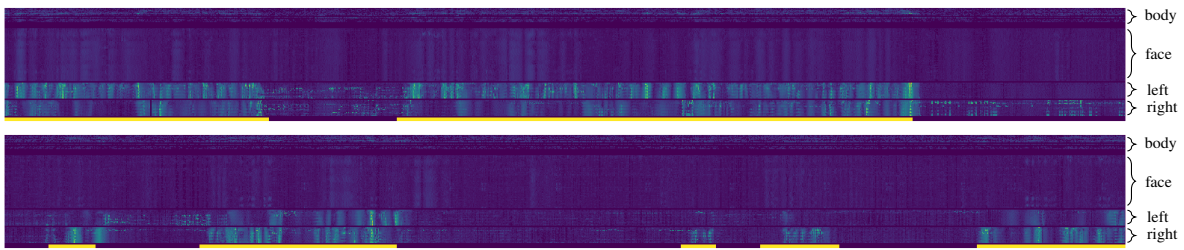
\begin{figure*}[hb]
    \begin{subfigure}{\linewidth}
        \begin{tikzpicture}
		% include the image
		\node[anchor=south east, inner sep=0] (image) at (0,0) {
            \includegraphics[width=0.95\linewidth,height=4em]{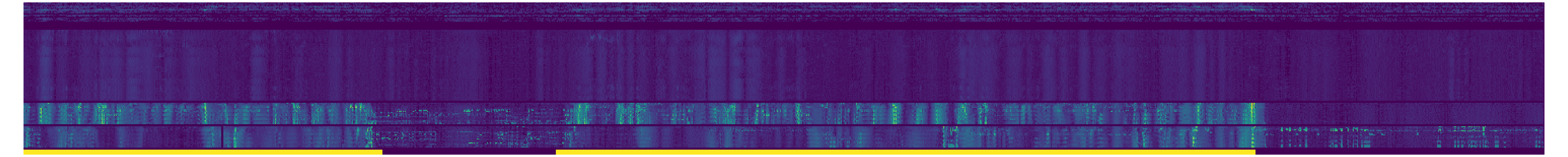}};
		
		% add the annotations
            \draw [decorate,decoration={brace,amplitude=3pt,mirror},yshift=7pt,xshift=-6pt]
		(0,1.07) -- (0,1.25) node [black,midway,xshift=0.44cm] {\tiny body };
		
		\draw [decorate,decoration={brace,amplitude=3pt,mirror},yshift=5pt,xshift=-6pt]
		(0,0.42) -- (0,1.06) node [black,midway,xshift=0.41cm] {\tiny face};
		
		\draw [decorate,decoration={brace,amplitude=3pt,mirror},yshift=6pt,xshift=-6pt]
		(0,0.14) -- (0,0.32) node [black,midway,xshift=0.4cm] {\tiny left};
		
		\draw [decorate,decoration={brace,amplitude=3pt,mirror},yshift=1,xshift=-6pt]
		(0,0.07) -- (0,0.25) node [black,midway,xshift=0.46cm] {\tiny right};
	\end{tikzpicture}
        
        \label{fig:rep:1}
    \end{subfigure}
    \begin{subfigure}{\linewidth}
	\begin{tikzpicture}
		% include the image
		\node[anchor=south east, inner sep=0] (image) at (0,0) {
            \includegraphics[width=0.95\linewidth,height=4em]{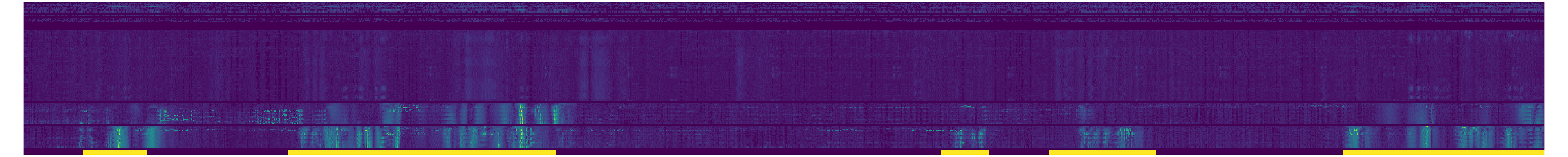}};
		
		% add the annotations
            \draw [decorate,decoration={brace,amplitude=3pt,mirror},yshift=7pt,xshift=-6pt]
		(0,1.07) -- (0,1.25) node [black,midway,xshift=0.44cm] {\tiny body };
		
		\draw [decorate,decoration={brace,amplitude=3pt,mirror},yshift=5pt,xshift=-6pt]
		(0,0.42) -- (0,1.06) node [black,midway,xshift=0.41cm] {\tiny face};
		
		\draw [decorate,decoration={brace,amplitude=3pt,mirror},yshift=6pt,xshift=-6pt]
		(0,0.14) -- (0,0.32) node [black,midway,xshift=0.4cm] {\tiny left};
		
		\draw [decorate,decoration={brace,amplitude=3pt,mirror},yshift=1,xshift=-6pt]
		(0,0.07) -- (0,0.25) node [black,midway,xshift=0.46cm] {\tiny right};
	\end{tikzpicture}
        
        \label{fig:rep:2}
    \end{subfigure}
    \caption{Optical flow for a conversation between two signers (signer 1 top, signer 2 bottom).  
    The x-axis is the progression across 30 seconds. %of time, %1,500 frames over 30 seconds in total. 
    The yellow marks the annotated phrase spans. (Source: \citet{detection:moryossef2020real})}
    \label{fig:phrase-boundaries}
\end{figure*}

\subsection{Boundary Modeling}\label{sec:motivation-bio}

When examining the nature of sign language expressions, we note that the utterances are typically signed in a continuous flow, with minimal to no pauses between individual signs. This continuity is particularly evident when dealing with lower frame rates. This continuous nature presents a significant difference from \emph{text} where specific punctuation marks serve as indicators of phrase boundaries, and a semi-closed set of tokens represent the \emph{words}.

Given these characteristics, directly applying conventional segmentation or sign language detection models---that is, utilizing IO tagging in a manner similar to image or audio segmentation models---may not yield the optimal solution, particularly at the sign level. Such models often fail to precisely identify the boundaries between signs.

A promising alternative is the Beginning-Inside-Outside (BIO) tagging \citep{ramshaw-marcus-1995-text}. BIO tagging was originally used for named entity recognition, but its application extends to any sequence chunking task beyond the text modality. In the context of sign language, BIO tagging provides a more refined model for discerning boundaries between signs and phrases, thus significantly improving segmentation performance (Figure \ref{fig:bio-example}).

\begin{figure}[ht]
\centering
\includegraphics[width=\linewidth]{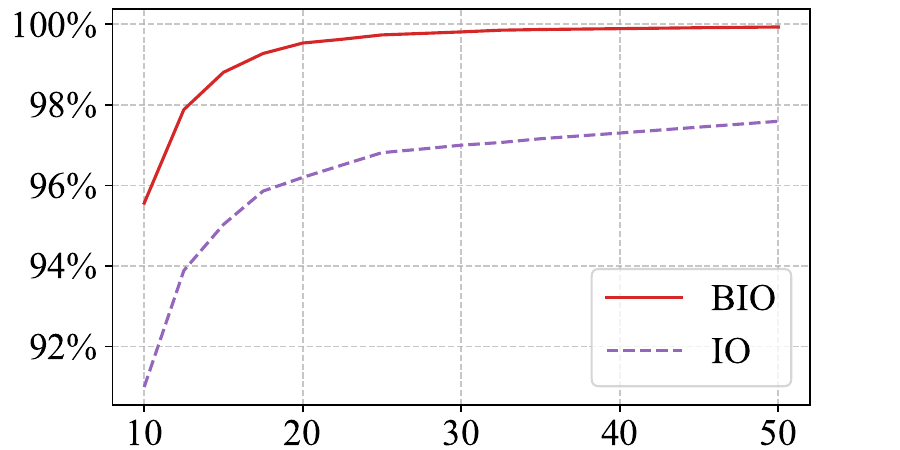}
\caption{Reproduced sign segments in the Public DGS Corpus by BIO and IO tagging at various frame rates. $99.7\%$ of segments reproduced at 25fps by BIO tagging.}
\label{fig:tagging-comparison}
\end{figure}

To test the viability of the BIO tagging approach in comparison with the IO tagging, we conducted an experiment on the Public DGS Corpus. 
The entire corpus was transformed to various frame rates and the sign segments were converted to frames using either BIO or IO tagging, then decoded back into sign segments.
Figure \ref{fig:tagging-comparison} illustrates the results of this comparison. Note that the IO tagging was unable to reproduce the same number of segments as the BIO tagging on the gold data. This underscores the importance of BIO tagging in identifying sign and phrase boundaries.

\subsection{Phrase Boundaries}\label{sec:motivation-optical-flow}

Linguistic research has shown that prosody is a reliable predictor of phrase boundaries in signed languages \citep{sandler2010prosody, ormel2012prosodic}. 
We observe that this is also the case in the Public DGS Corpus dataset used in our experiments. 
To illustrate this, we model pauses and movement using optical flow directly on the poses as proposed by \citet{detection:moryossef2020real}. Figure \ref{fig:phrase-boundaries} demonstrates that a change in motion signifies the presence of a pause, which, in turn, indicates a phrase boundary.

\subsection{Sign Boundaries}\label{sec:motivation-hand-shapes}

We observe that signs generally utilize a limited number of hand shapes, with the majority of signs utilizing a maximum of two hand shapes. For example, linguistically annotated datasets, such as ASL-LEX \cite{sehyr2021asl} and ASLLVD \cite{neidle2012challenges}, only record one initial hand shape per hand and one final hand shape. 
\citet[p.\,87]{Mandel1981PhonotacticsAM} argued that there can only be one set of selected fingers per sign, constraining the number of handshapes in signs. This limitation is referred to as the \emph{Selected Fingers Constraint}. 
And indeed, \citet{Sandler2008TheSI} find that the optimal form of a sign is monosyllabic, and that handshape change is organized by the syllable unit.

\begin{figure}[htbp]
\centering
\includegraphics[width=\linewidth]{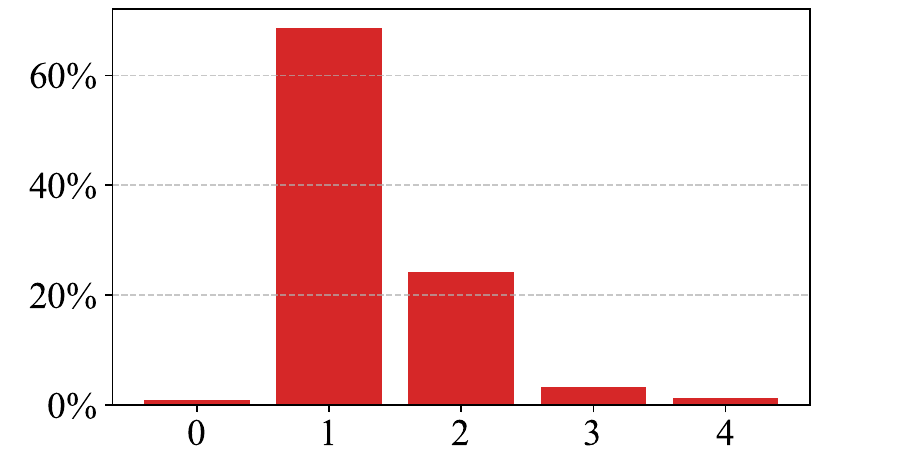}
\caption{Number of hand shapes per sign in SignBank.}
\label{fig:sign-boundaries}
\end{figure}

To illustrate this constraint empirically, we show a histogram of hand shapes per sign in SignBank\footnote{\url{https://signbank.org/signpuddle2.0/}} for $705,151$ signs in Figure \ref{fig:sign-boundaries}.

Additionally, we found that a change in the dominant hand shape often signals the presence of a sign boundary. Specifically, out of $27,658$ sentences, comprising $354,955$ pairs of consecutive signs, only $17.38\%$ of consecutive signs share the same base hand shape\footnote{It is important to note that this percentage is inflated, as it may encompass overlaps across the dominant and non-dominant hands, which were not separated for this analysis.}.
Based on these observations, we propose using 3D hand normalization as an indicative cue for hand shapes to assist in detecting sign boundaries. We hypothesize that performing 3D hand normalization makes it easier for the model to extract the hand shape.
We expand on this process and show examples in Appendix \ref{appendix:hand-analysis}.

\section{Experimental Setup}\label{sec:setup}
In this section, we describe the experimental setup used to evaluate our linguistically motivated approach for sign language segmentation.
This includes a description of the Public DGS Corpus dataset used in the study, the methodology employed to perform sign and phrase segmentation, and the evaluation metrics used to measure the performance of the proposed approach.

\subsection{Dataset}\label{sec:dataset}

The Public DGS Corpus \cite{dataset:prillwitz2008dgs, dataset:hanke-etal-2020-extending} is a distinctive sign language dataset that includes both accurate sign-level annotation from continuous signing, and well-aligned phrase-level translation in spoken language.

The corpus comprises 404 documents / 714 videos\footnote{The number of videos is nearly double the number of documents because each document typically includes two signers, each of whom produces one video for segmentation.} with an average duration of 7.55 minutes, featuring either one signer or two signers, at 50 fps. 
Most of these videos feature gloss transcriptions and spoken language translations (German and English), except for the ones in the ``Joke'' category, which are not annotated and thus excluded from our model\footnote{We also exclude documents with missing annotations. 
$id \in \{1289910, 1245887, 1289868, 1246064, 1584617\}$}.
The translations are comprised of full spoken language paragraphs, sentences, or phrases (i.e., independent/main clauses).

Each gloss span is considered a gold sign segment, following a tight annotation scheme \citep{Hanke2012WhereDA}.
Phrase segments are identified by examining every translation, with the segment assumed to span from the start of its first sign to the end of its last sign, correcting imprecise annotation.

This corpus is enriched with full-body pose estimations from OpenPose \cite{pose:cao2018openpose,schulderopenpose} and Mediapipe Holistic \cite{mediapipe2020holistic}.
We use the \emph{3.0.0-uzh-document} split from \citet{zhang2023sltunet}. After filtering the unannotated data, we are left with 296 documents / 583 videos for training, 6 / 12 for validation, and 9 / 17 for testing. The mean number of signs and phrases in a video from the training set is 613 and 111, respectively.

\subsection{Methodology}\label{sec:method}

Our proposed approach for sign language segmentation is based on the following steps:

\begin{enumerate}
    \item \textbf{Pose Estimation} 
    Given a video, we first adjust it to 25 fps and estimate body poses using the MediaPipe Holistic pose estimation system. We do not use OpenPose because it lacks a $Z$-axis, which prevents 3D rotation used for hand normalization. The shape of a pose is represented as $(\text{frames} \times \text{keypoints} \times \text{axes})$.

    \item \textbf{Pose Normalization} 
    To generalize over video resolution and distance from the camera, we normalize each of these poses such that the mean distance between the shoulders of each person equals $1$, and the mid-point is at $(0,0)$ \citep{Celebi2013GestureRU}. We also remove the legs since they are less relevant to signing.
    
    \item \textbf{Optical Flow} 
    We follow the equation in \citet[Equation 1]{detection:moryossef2020real}.
    
    \item \textbf{3D Hand Normalization} 
    We rotate and scale each hand to ensure that the same hand shape is represented in a consistent manner across different frames. We rotate the 21 $XYZ$ keypoints of the hand so that the back of the hand lies on the $XY$ plane, we then rotate the hand so that the metacarpal bone of the middle finger lies on the $Y$-axis, and finally, we scale the hand such that the bone is of constant length.
    Visualizations are presented in Appendix \ref{appendix:hand-analysis}.
    
    \item \textbf{Sequence Encoder}
    For every frame, the pose is first flattened and projected into a standard dimension ($256$), then fed through an LSTM encoder \citep{hochreiter1997long}. 
    
    \item \textbf{BIO Tagging} 
    On top of the encoder, we place two BIO classification heads for sign and phrase independently. \emph{B} denotes the beginning of a sign or phrase, \emph{I} denotes the middle of a sign or phrase, and \emph{O} denotes being outside a sign or phrase.
    Our cross-entropy loss is proportionally weighted in favor of \emph{B} as it is a \emph{rare} label\footnote{B:I:O is about 1:5:18 for signs and 1:58:77 for phrases.} compared to \emph{I} and \emph{O}.
    
    \item \textbf{Greedy Segment Decoding} 
    To decode the frame-level BIO predictions into sign/phrase segments, we define a segment to start with the first frame possessing a \emph{B} probability greater than a predetermined threshold (defaulted at $0.5$). The segment concludes with the first frame among the subsequent frames, having either a \emph{B} or \emph{O} probability exceeding the threshold. We provide the pseudocode of the decoding algorithm in Appendix \ref{appendix:greedy-decoding}.
\end{enumerate}
 
\subsection{Experiments}\label{sec:experiments}
Our approach is evaluated through a series of six sets of experiments. Each set is repeated three times with varying random seeds. Preliminary experiments were conducted to inform the selection of hyperparameters and features, the details of which can be found in Table \ref{tab:preliminary} in Appendix \ref{appendix:extended-results}. Model selection is based on validation metrics.

\begin{table*}[htbp]
% \small
\centering
\resizebox{\textwidth}{!}{%
\begin{tabular}{llccc|ccc|cc}
\toprule
 & & \multicolumn{3}{c}{\textbf{Sign}} & \multicolumn{3}{c}{\textbf{Phrase}} & \multicolumn{2}{c}{\textbf{Efficiency}} \\
\cmidrule(lr){3-5} \cmidrule(lr){6-8} \cmidrule(lr){9-10}
\multicolumn{2}{l}{\textbf{Experiment}} & \textbf{F1} & \textbf{IoU} & \textbf{\%} & \textbf{F1} & \textbf{IoU} & \textbf{\%} & \textbf{\#Params} & \textbf{Time} \\
\midrule

\textbf{E0} & \textbf{\citet{detection:moryossef2020real}} & --- & $0.46$ & $1.09$ & --- & $0.70$ & $\textbf{1.00}$ & \textbf{102K} & \textbf{0:50:17}\\
\midrule
\textbf{E1} & \textbf{Baseline} & $0.56$ & $0.66$ & $0.91$ & $0.59$ & $0.80$ & $2.50$ & 454K & 1:01:50\\
\textbf{E2} & \textbf{E1 + Face} & $0.53$ & $0.58$ & $0.64$ & $0.57$ & $0.76$ & $1.87$ & 552K & 1:50:31\\
\textbf{E3} & \textbf{E1 + Optical Flow} & $0.58$ & $0.62$ & $1.12$ & $0.60$ & $0.82$ & $3.19$ & 473K & 1:20:17\\
\textbf{E4} & \textbf{E3 + Hand Norm} & $0.56$ & $0.61$ & $1.07$ & $0.60$ & $0.80$ & $3.24$ & 516K & 1:30:59\\
\midrule
\textbf{E1s} & \textbf{E1 + Depth=4} & $\textbf{0.63}$ & $\textbf{0.69}$ & $1.11$ & $\textbf{0.65}$ & $0.82$ & $1.63$ & 1.6M & 4:08:48\\
\textbf{E2s} & \textbf{E2 + Depth=4} & $0.62$ & $\textbf{0.69}$ & $1.07$ & $0.63$ & $0.84$ & $2.68$ & 1.7M & 3:14:03\\
\textbf{E3s} & \textbf{E3 + Depth=4} & $0.60$ & $0.63$ & $1.13$ & $0.64$ & $0.80$ & $1.53$ & 1.7M & 4:08:30\\
\textbf{E4s} & \textbf{E4 + Depth=4} & $0.59$ & $0.63$ & $1.13$ & $0.62$ & $0.79$ & $1.43$ & 1.7M & 4:35:29\\
\midrule
\textbf{E1s*} & \textbf{E1s + Tuned Decoding} & --- & \textbf{0.69} & \textbf{1.03} & --- & \textbf{0.85} & 1.02 & --- & ---\\
\textbf{E4s*} & \textbf{E4s + Tuned Decoding} & --- & 0.63 & 1.06 & --- & 0.79 & 1.12 & --- & ---\\
\midrule
\textbf{E5} & \textbf{E4s + Autoregressive} & $0.45$ & $0.47$ & $0.88$ & $0.52$ & $0.63$ & $2.72$ & 1.3M & \textasciitilde3 days\\

\bottomrule
\end{tabular}
}
\caption{Mean test evaluation metrics for our experiments. The best score of each column is in bold and a star (*) denotes further optimization of the decoding algorithm without changing the model (only affects \emph{IoU} and \emph{\%}). Table \ref{tab:results-full} in Appendix \ref{appendix:extended-results} contains a complete report including validation metrics and standard deviation of all experiments.}
\label{tab:results}
\end{table*}

\begin{enumerate}

\item \textbf{E0: IO Tagger} 
We re-implemented and reproduced\footnote{The initial implementation uses OpenPose \citep{pose:cao2018openpose}, at 50 fps. Preliminary experiments reveal that these differences do not significantly influence the results.} the sign language detection model proposed by \citet{detection:moryossef2020real}, in PyTorch \citep{NEURIPS2019_9015} as a naive baseline. This model processes optical flow as input and outputs \emph{I} (is signing) and \emph{O} (not signing) tags.

\item \textbf{E1: Bidirectional BIO Tagger} 
We replace the IO tagging heads in \emph{E0} with BIO heads to form our baseline. Our preliminary experiments indicate that inputting only the 75 hand and body keypoints and making the LSTM layer bidirectional yields optimal results. % with an option to increase the number of LSTM layers from 1 to 4.

\item \textbf{E2: Adding Reduced Face Keypoints} 
Although the 75 hand and body keypoints serve as an efficient minimal set for sign language detection/segmentation models, we investigate the impact of other nonmanual sign language articulators, namely, the face. We introduce a reduced set of 128 face keypoints that signify the signer's \emph{face contour}\footnote{We reduce the dense \emph{FACE\_LANDMARKS} in Mediapipe Holistic to the contour keypoints according to the variable \emph{mediapipe.solutions.holistic.FACEMESH\_CONTOURS}.}.

\item \textbf{E3: Adding Optical Flow} 
At every time step $t$ we append the optical flow between $t$ and $t-1$ to the current pose frame as an additional dimension after the $XYZ$ axes.

\item \textbf{E4: Adding 3D Hand Normalization} 
At every time step, we normalize the hand poses and concatenate them to the current pose.

\item \textbf{E5: Autoregressive Encoder} 
We replace the encoder with the one proposed by \citet{jiang2023automatic} for the detection and classification of great ape calls from raw audio signals. Specifically, we add autoregressive connections between time steps to encourage consistent output labels. The logits at time step $t$ are concatenated to the input of the next time step, $t+1$. This modification is implemented bidirectionally by stacking two autoregressive encoders and adding their output up before the Softmax operation. However, this approach is inherently slow, as we have to fully wait for the previous time step predictions before we can feed them to the next time step.

\end{enumerate}

\subsection{Evaluation Metrics}
We evaluate the performance of our proposed approach for sign and phrase segmentation using the following metrics:

\begin{itemize}
    \item \textbf{Frame-level F1 Score} For each frame, we apply the \textit{argmax} operation to make a local prediction of the BIO class and calculate the macro-averaged per-class F1 score against the ground truth label. We use this frame-level metric during validation as the primary metric for model selection and early stopping, due to its independence from a potentially variable segment decoding algorithm (\S\ref{sec:tuning-decoding}).

    \item \textbf{Intersection over Union (IoU)} We compute the IoU between the ground truth segments and the predicted segments to measure the degree of overlap. Note that we do not perform a one-to-one mapping between the two using techniques like DTW. Instead, we calculate the total IoU based on all segments in a video.
    
    \item \textbf{Percentage of Segments (\%)} 
    To complement IoU, we introduce the percentage of segments to compare the number of segments predicted by the model with the ground truth annotations. It is computed as follows: $\frac{\# \text{predicted segments}}{\# \text{ground truth segments}}$. The optimal value is 1.

    \item \textbf{Efficiency} We measure the efficiency of each model by the number of parameters and the training time of the model on a Tesla V100-SXM2-32GB GPU for 100 epochs\footnote{Exceptionally the autoregressive models in \textit{E5} were trained on an NVIDIA A100-SXM4-80GB GPUA100 which doubles the training speed of V100, still the training is slow.}.
\end{itemize}

\section{Results and Discussion}\label{sec:results}

We report the mean test evaluation metrics for our experiments in Table \ref{tab:results}.
We do not report F1 Score for \emph{E0} since it has a different number of classes and is thus incomparable.
Comparing \emph{E1} to \emph{E0}, we note that the model's bidirectionality, the use of poses, and BIO tagging indeed help outperform the model from previous work where only optical flow and IO tagging are used. While \emph{E1} predicts an excessive number of phrase segments, the IoUs for signs and phrases are both higher. 

Adding face keypoints (\emph{E2}) makes the model worse, while including optical flow (\emph{E3}) improves the F1 scores. For phrase segmentation, including optical flow increases IoU, but over-segments phrases by more than 300\%, which further exaggerates the issue in \emph{E1}. Including hand normalization (\emph{E4}) on top of \emph{E3} slightly deteriorates the quality of both sign and phrase segmentation.

From the non-exhaustive hyperparameter search in the preliminary experiments (Table \ref{tab:preliminary}), we examined different hidden state sizes ($64$, $128$, $256$, $512$, $1024$) and a range of $1$ to $8$ LSTM layers, and conclude that a hidden size of $256$ and $4$ layers with $1e-3$ learning rate are optimal for \emph{E1}, which lead to \emph{E1s}.
We repeat the setup of \emph{E2}, \emph{E3}, and \emph{E4} with these refined hyper-parameters, and show that all of them outperform their counterparts, notably in that they ease the phrase over-segmentation problem. 

In \emph{E2s}, we reaffirmed that adding face keypoints does not yield beneficial results, so we exclude face in future experiments. Although the face is an essential component to understanding sign language expressions and does play some role in sign and phrase level segmentation, we believe that the 128 face contour points are too dense for the model to learn useful information compared to the 75 body points, and may instead confuse the model.

In addition, the benefits of explicitly including optical flow (\emph{E3s}) fade away with the increased model depth and we speculate that now the model might be able to learn the optical flow features by itself. 
Surprisingly, while adding hand normalization (\emph{E4s}) still slightly worsens the overall results, it has the best phrase percentage.

From \emph{E4s} we proceeded with the training of \emph{E5}, an autoregressive model. Unexpectedly, counter to our intuition and previous work, \emph{E5} underachieves our baseline across all evaluation metrics\footnote{\emph{E5} should have more parameters than \emph{E4s}, but because of an implementation bug, each LSTM layer has half the parameters. Based on the current results, we assume that autoregressive connections (even with more parameters) will not improve our models.}.

\subsection{Challenges with 3D Hand Normalization}

While the use of 3D hand normalization is well-justified in \S\ref{sec:motivation-hand-shapes}, we believe it does not help the model due to poor depth estimation quality, as further corroborated by recent research from \citet{de2023towards}. Therefore, we consider it a negative result, showing the deficiencies in the 3D pose estimation system. The evaluation metrics we propose in Appendix \ref{appendix:hand-analysis} could help identify better pose estimation models for this use case.

\subsection{Tuning the Segment Decoding Algorithm}\label{sec:tuning-decoding}

We selected \emph{E1s} and \emph{E4s} to further explore the segment decoding algorithm. As detailed in \S\ref{sec:method} and Appendix \ref{appendix:greedy-decoding}, the decoding algorithm has two tunable parameters, $threshold_b$ and $threshold_o$. We conducted a grid search with these parameters, using values from 10 to 90 in increments of 10. We additionally experimented with a variation of the algorithm that conditions on the most likely class by \emph{argmax} instead of fixed threshold values, which turned out similar to the default version.

We only measured the results using IoU and the percentage of segments at validation time since the F1 scores remain consistent in this case. 
% We separately identified the best parameters for sign ($threshold_b=30, threshold_o=40$) and phrase ($threshold_b=70, threshold_o=90$) segmentation and reported the test results under \emph{E4s*}.
For sign segmentation, we found using $threshold_b=60$ and $threshold_o=40/50/60$ yields slightly better results than the default setting ($50$ for both).
For phrase segmentation, we identified that higher threshold values ($threshold_b=90, threshold_o=90$ for \textit{E1s} and $threshold_b=80, threshold_o=80/90$ for \textit{E4s}) improve on the default significantly, especially on the percentage metric.
We report the test results under \emph{E1s*} and \emph{E4s*}, respectively.

\begin{figure}
\centering
\includegraphics[width=\linewidth]{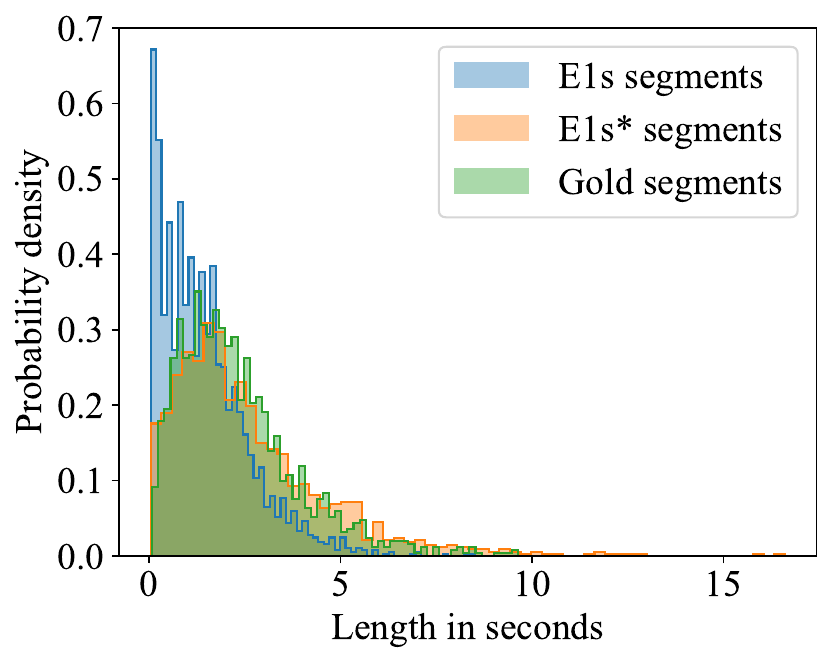}
\caption{Probability density of phrase segment lengths.} 
\label{fig:segment-distribution}
\end{figure}

Despite formulating a single model, we underline a separate sign/phrase model selection process to archive the best segmentation results.
Figure \ref{fig:segment-distribution} illustrates how higher threshold values reduce the number of predicted segments and skew the distribution of predicted phrase segments towards longer ones in \emph{E1s}/\emph{E1s*}. As \citet{segmentation:bull2020automatic} suggest, advanced priors could also be integrated into the decoding algorithm.

\subsection{Comparison to Previous Work}\label{sec:comparison}

We re-implemented and re-purposed the sign language detection model introduced in \citet{detection:moryossef2020real} for our segmentation task as a baseline since their work is the state-of-the-art and the only comparable model designed for the Public DGS Corpus dataset. As a result, we show the necessity of replacing IO tagging with BIO tagging to tackle the subtle differences between the two tasks.

For \emph{phrase} segmentation, we compare to \citet{segmentation:bull2020automatic}.
We note that our definition of sign language phrases (spanning from the start of its first sign to the end of its last sign) is tighter than the subtitle units used in their paper and that we use different training datasets of different languages and domains.
Nevertheless, we implemented some of their frame-level metrics and show the results in Table \ref{tab:comparison} on both the Public DGS Corpus and the MEDIAPI-SKEL dataset \citep{bull-etal-2020-mediapi} in French Sign Language (LSF).
We report both zero-shot out-of-domain results\footnote{The zero-shot results are not directly comparable to theirs due to different datasets and labeling approaches.} and the results of our models trained specifically on their dataset without the spatio-temporal graph convolutional network (ST-GCN) \citep{yan2018spatial} used in their work for pose encoding.
%the results are not directly comparable.%\footnote{As of writing, we have not yet been granted access to the MEDIAPI-SKEL dataset \citep{bull-etal-2020-mediapi}. Once access is granted, a more fair comparison can be made.}.

\begin{table}[htbp]
\centering
\resizebox{\linewidth}{!}{%
% \begin{tabular}{ll|cccc|c}
% \toprule
% \textbf{Data} & \textbf{Model} & \textbf{ROC-AUC} & \textbf{Precision} & \textbf{Recall} & \textbf{F1} & \textbf{F1-M}\\
% \midrule
% \multirow{6}{*}{LSF} & \textbf{full (theirs)} & 0.87 & 0.50 & 0.75 & 0.60 & ---\\
% & \textbf{body (theirs)} & 0.87 & 0.56 & 0.71 & 0.63 & ---\\
% % full (theirs) & 0.8723 & 0.5023 & 0.7510 & 0.6019 & ---\\
% % body (theirs) & 0.8704 & 0.5616 & 0.7122 & 0.6280 & ---\\
% \cmidrule{2-7}
% & \textbf{E1s (ours, trained)} & 0.87 & & & & 0.49\\
% & \textbf{E4s (ours, trained)} & 0.87 & & & & 0.51\\
% \cmidrule{2-7}
% & \textbf{E1s (ours, zero-shot)} & 0.71 & 0.36 & 0.46 & 0.37 & 0.41\\
% & \textbf{E4s (ours, zero-shot)} & 0.76 & 0.41 & 0.58 & 0.46 & 0.44\\
% \midrule
% \multirow{2}{*}{DGS} & \textbf{E1s (ours)} & 0.91 & 0.84 & 0.74 & 0.78 & 0.65\\
% & \textbf{E4s (ours)} & 0.90 & 0.83 & 0.74 & 0.78 & 0.62\\
\begin{tabular}{ll|cc}
\toprule
\textbf{Data} & \textbf{Model} & \textbf{ROC-AUC} & \textbf{F1-M}\\
\midrule
\multirow{6}{*}{LSF} & \textbf{full (theirs)} & 0.87 &  ---\\
& \textbf{body (theirs)} & 0.87 & ---\\
% full (theirs) & 0.8723 & 0.5023 & 0.7510 & 0.6019 & ---\\
% body (theirs) & 0.8704 & 0.5616 & 0.7122 & 0.6280 & ---\\
\cmidrule{2-4}
& \textbf{E1s (ours, zero-shot)} & 0.71 & 0.41\\
& \textbf{E4s (ours, zero-shot)} & 0.76 & 0.44\\
\cmidrule{2-4}
& \textbf{E1s (ours, trained)} & 0.87 & 0.49\\
& \textbf{E4s (ours, trained)} & 0.87 & 0.51\\
\midrule
\multirow{2}{*}{DGS} & \textbf{E1s (ours)} & 0.91 & 0.65\\
& \textbf{E4s (ours)} & 0.90 & 0.62\\
\bottomrule
\end{tabular}
}
% \caption{Evaluation metrics used in \citet{segmentation:bull2020automatic}. All their metrics are applied exclusively on the \emph{O}-tag. 
% For comparison \emph{F1-M} denotes the macro-averaged per-class F1 used in this work across all tags. 
% The first two rows are the best results taken from Table 1 in their paper. The next two rows represent how our models perform on their data in a zero-shot setting, and the last two rows represent how our models perform on our data.}
\caption{Evaluation metrics used in \citet{segmentation:bull2020automatic}. \emph{ROC-AUC} is applied exclusively on the \emph{O}-tag. 
For comparison \emph{F1-M} denotes the macro-averaged per-class F1 used in this work across all tags. 
The first two rows are the best results taken from Table 1 in their paper. The next four rows represent how our models perform on their data in a zero-shot setting, and in a supervised setting, and the last two rows represent how our models perform on our data.}
\label{tab:comparison}
\end{table}

For \emph{sign} segmentation, we do not compare to \citet{segmentation:renz2021signa, segmentation:renz2021signb} due to different datasets and the difficulty in reproducing their segment-level evaluation metrics. The latter depends on the decoding algorithm and a way to match the gold and predicted segments, both of which are variable.

\section{Conclusions}\label{sec:conclusion}

% 1) this work is on sign language segmentation, we are the first to formulate the segmentation of individual signs and larger sign phrases as a joint problem

This work focuses on the automatic segmentation of signed languages. We are the first to formulate the segmentation of individual signs and larger sign phrases as a joint problem.
 
% 2) Our approach is linguistically motivated. We report a series of analyses of sign language corpora that informs our experiments.
% - continuous signing: BIO instead of IO
% - optical flow
% - a careful analysis of the number and properties of handshapes in corpora

We propose a series of improvements over previous work, linguistically motivated by careful analyses of sign language corpora. 
Recognizing that sign language utterances are typically continuous with minimal pauses, we opted for a BIO tagging scheme over IO tagging. Furthermore, leveraging the fact that phrase boundaries are marked by prosodic cues, we introduce optical flow features as a proxy for prosodic processes. 
Finally, since signs typically employ a limited number of hand shapes, to make it easier for the model to understand handshapes, we attempt 3D hand normalization.

% 3) the experiments we performed, on which corpus

% 4) main findings:
% - BIO tagging is indeed better
% - optical flow to encode prosody also improves segmentation accuracy
% - tuning the hyperparameters for decoding improves the base model?

Our experiments conducted on the Public DGS Corpus confirmed the efficacy of these modifications for segmentation quality. 
By comparing to previous work in a zero-shot setting, we demonstrate that our models generalize across signed languages and domains and that including linguistically motivated cues leads to a more robust model in this context. 

Finally, we envision that the proposed model has applications in real-world data collection for signed languages.
Furthermore, a similar segmentation approach could be leveraged in various other fields such as co-speech gesture recognition \citep{moryossef2023addressing} and action segmentation \citep{tang2019coin}.

% \input{figures/object-detection/object-detection}

% \clearpage

% Mandatory section
\section*{Limitations}\label{sec:limitation}

\subsection*{Pose Estimation}
In this work, we employ the MediaPipe Holistic pose estimation system \citep{mediapipe2020holistic}. There is a possibility that this system exhibits bias towards certain protected classes (such as gender or race), underperforming in instances with specific skin tones or lower video quality. Thus, we cannot attest to how our system would perform under real-world conditions, given that the videos utilized in our research are generated in a controlled studio environment, primarily featuring white participants.

\subsection*{Encoding of Long Sequences}

In this study, we encode sequences of frames that are significantly longer than the typical 512 frames often seen in models employing Transformers \citep{vaswani2017attention}. Numerous techniques, ranging from basic temporal pooling/downsampling to more advanced methods such as a video/pose encoder that aggregates local frames into higher-level `tokens' \citep{segmentation:renz2021signa}, graph convolutional networks \citep{segmentation:bull2020automatic}, and self-supervised representations \citep{baevski2020wav2vec}, can alleviate length constraints, facilitate the use of Transformers, and potentially improve the outcomes. Moreover, a hierarchical method like the Swin Transformer \citep{liu2021swin} could be applicable.

\subsection*{Limitations of Autoregressive LSTMs}
In this paper, we replicated the autoregressive LSTM implementation originally proposed by \citet{jiang2023automatic}.
Our experiments revealed that this implementation exhibits significant slowness, which prevented us from performing further experimentation. 
In contrast, other LSTM implementations employed in this project have undergone extensive optimization \citep{Appleyard2016}, including techniques like combining general matrix multiplication operations (GEMMs), parallelizing independent operations, fusing kernels, rearranging matrices, and implementing various optimizations for models with multiple layers (which are not necessarily applicable here).

A comparison of CPU-based performance demonstrates that our implementation is x6.4 times slower. 
Theoretically, the number of operations performed by the autoregressive LSTM is equivalent to that of a regular LSTM. However, while the normal LSTM benefits from concurrency based on the number of layers, we do not have that luxury. 
The optimization of recurrent neural networks (RNNs) \citep{Que2020OptimizingRR,Que2021AcceleratingRN,Que2022RemarnAR} remains an ongoing area of research. If proven effective in other domains, we strongly advocate for efforts to optimize the performance of this type of network.

\subsection*{Interference Between Sign and Phrase Models}

In our model, we share the encoder for both the sign and phrase segmentation models, with a shallow linear layer for the BIO tag prediction associated with each task. 
It remains uncertain whether these two tasks interfere with or enhance each other. 
An ablation study (not presented in this work) involving separate modeling is necessary to obtain greater insight into this matter.

\subsection*{Noisy Training Objective}

Although the annotations utilized in this study are of expert level, the determination of precise sign \citep{Hanke2012WhereDA} and phrase boundaries remains a challenging task, even for experts. Training the model on these annotated boundaries might introduce excessive noise. 
A similar issue was observed in classification-based pose estimation \citep{pose:cao2018openpose}. The task of annotating the exact anatomical centers of joints proves to be nearly impossible, leading to a high degree of noise when predicting joint position as a 1-hot classification task. The solution proposed in this previous work was to distribute a Gaussian around the annotated location of each joint. This approach allows the joint's center to overlap with some probability mass, thereby reducing the noise for the model. A similar solution could be applied in our context. Instead of predicting a strict 0 or 1 class probability, we could distribute a Gaussian around the boundary.

\subsection*{Naive Segment Decoding}

We recognize that the frame-level greedy decoding strategy implemented in our study may not be optimal. 
Previous research in audio segmentation \citep{venkatesh2022you} employed a You Only Look Once (YOLO; \citet{Redmon2015YouOL}) decoding scheme to predict segment boundaries and classes.
We propose using a similar prediction atop a given representation, such as the LSTM output or classification logits of an already trained network. Differing from traditional object detection tasks, this process is simplified due to the absence of a $Y$ axis and non-overlapping segments. In this scenario, the network predicts the segment boundaries using regression, thereby avoiding the class imbalance issue of the BIO tagging. We anticipate this to yield more accurate sign language segmentation.

\subsection*{Lack of Transcription}

Speech segmentation is a close task to our sign language segmentation task on videos. In addition to relying on prosodic cues from audio,  the former could benefit from automatic speech transcription systems, either in terms of surrogating the task to text-level segmentation and punctuation \citep{cho2015punctuation}, or gaining additional training data from automatic speech recognition / spoken language translation \citep{tsiamas22_interspeech}.

However, for signed languages, there is neither a standardized and widely used written form nor a reliable transcription procedure into some potential writing systems like SignWriting \citep{writing:sutton1990lessons}, HamNoSys \citep{writing:prillwitz1990hamburg}, and glosses \citep{Johnston2008FromAT}. 
Transcription/recognition and segmentation tasks need to be solved simultaneously, so we envision that a multi-task setting helps.
Sign spotting, the localization of a specific sign in continuous signing, is a simplification of the segmentation and recognition problem in a closed-vocabulary setting \citep{Wong2022HierarchicalIF,Varol2022ScalingUS}. It can be used to find candidate boundaries for some signs, but not all.

\section*{Acknowledgements}

This work was funded by the EU Horizon 2020 project EASIER (grant agreement no. 101016982), the Swiss Innovation Agency  (Innosuisse) flagship IICT (PFFS-21-47) and the EU Horizon 2020 project iEXTRACT (grant agreement no. 802774). We also thank Rico Sennrich and Chantal Amrhein for their suggestions.

\bibliography{anthology,custom,background}

\begin{thebibliography}{58}
\expandafter\ifx\csname natexlab\endcsname\relax\def\natexlab#1{#1}\fi

\bibitem[{Appleyard(2016)}]{Appleyard2016}
Jeremy Appleyard. 2016.
\newblock \href
  {https://developer.nvidia.com/blog/optimizing-recurrent-neural-networks-cudnn-5/}
  {Optimizing recurrent neural networks in {cuDNN} 5}.
\newblock Accessed: 2023-06-09.

\bibitem[{Baevski et~al.(2020)Baevski, Zhou, Mohamed, and
  Auli}]{baevski2020wav2vec}
Alexei Baevski, Yuhao Zhou, Abdelrahman Mohamed, and Michael Auli. 2020.
\newblock wav2vec 2.0: A framework for self-supervised learning of speech
  representations.
\newblock \emph{Advances in neural information processing systems},
  33:12449--12460.

\bibitem[{Borg and Camilleri(2019)}]{detection:borg2019sign}
Mark Borg and Kenneth~P Camilleri. 2019.
\newblock Sign language detection "in the wild" with recurrent neural networks.
\newblock In \emph{ICASSP 2019-2019 IEEE International Conference on Acoustics,
  Speech and Signal Processing (ICASSP)}, pages 1637--1641. IEEE.

\bibitem[{Bull et~al.(2021)Bull, Afouras, Varol, Albanie, Momeni, and
  Zisserman}]{segmentation:bull2021aligning}
Hannah Bull, Triantafyllos Afouras, G{\"u}l Varol, Samuel Albanie, Liliane
  Momeni, and Andrew Zisserman. 2021.
\newblock Aligning subtitles in sign language videos.
\newblock In \emph{Proceedings of the IEEE/CVF International Conference on
  Computer Vision}, pages 11552--11561.

\bibitem[{Bull et~al.(2020{\natexlab{a}})Bull, Braffort, and
  Gouiff{\`e}s}]{bull-etal-2020-mediapi}
Hannah Bull, Annelies Braffort, and Mich{\`e}le Gouiff{\`e}s.
  2020{\natexlab{a}}.
\newblock \href {https://aclanthology.org/2020.lrec-1.743} {{MEDIAPI}-{SKEL} -
  a 2{D}-skeleton video database of {F}rench {S}ign {L}anguage with aligned
  {F}rench subtitles}.
\newblock In \emph{Proceedings of the Twelfth Language Resources and Evaluation
  Conference}, pages 6063--6068, Marseille, France. European Language Resources
  Association.

\bibitem[{Bull et~al.(2020{\natexlab{b}})Bull, Gouiff{\`e}s, and
  Braffort}]{segmentation:bull2020automatic}
Hannah Bull, Mich{\`e}le Gouiff{\`e}s, and Annelies Braffort.
  2020{\natexlab{b}}.
\newblock Automatic segmentation of sign language into subtitle-units.
\newblock In \emph{European Conference on Computer Vision}, pages 186--198.
  Springer.

\bibitem[{{Cao} et~al.(2019){Cao}, {Hidalgo Martinez}, {Simon}, {Wei}, and
  {Sheikh}}]{pose:cao2018openpose}
Z.~{Cao}, G.~{Hidalgo Martinez}, T.~{Simon}, S.~{Wei}, and Y.~A. {Sheikh}.
  2019.
\newblock {O}pen{P}ose: Realtime multi-person {2D} pose estimation using part
  affinity fields.
\newblock \emph{IEEE Transactions on Pattern Analysis and Machine
  Intelligence}.

\bibitem[{Celebi et~al.(2013)Celebi, Aydin, Temiz, and
  Arici}]{Celebi2013GestureRU}
Sait Celebi, Ali~Selman Aydin, Talha~Tarik Temiz, and Tarik Arici. 2013.
\newblock Gesture recognition using skeleton data with weighted dynamic time
  warping.
\newblock In \emph{International Conference on Computer Vision Theory and
  Applications}.

\bibitem[{Cho et~al.(2015)Cho, Niehues, Kilgour, and
  Waibel}]{cho2015punctuation}
Eunah Cho, Jan Niehues, Kevin Kilgour, and Alex Waibel. 2015.
\newblock Punctuation insertion for real-time spoken language translation.
\newblock In \emph{Proceedings of the 12th International Workshop on Spoken
  Language Translation: Papers}, pages 173--179.

\bibitem[{Cho et~al.(2014)Cho, van Merri{\"e}nboer, Gulcehre, Bahdanau,
  Bougares, Schwenk, and Bengio}]{cho2014learning}
Kyunghyun Cho, Bart van Merri{\"e}nboer, Caglar Gulcehre, Dzmitry Bahdanau,
  Fethi Bougares, Holger Schwenk, and Yoshua Bengio. 2014.
\newblock \href {https://doi.org/10.3115/v1/D14-1179} {Learning phrase
  representations using {RNN} encoder{--}decoder for statistical machine
  translation}.
\newblock In \emph{Proceedings of the 2014 Conference on Empirical Methods in
  Natural Language Processing ({EMNLP})}, pages 1724--1734, Doha, Qatar.
  Association for Computational Linguistics.

\bibitem[{De~Coster et~al.(2023)De~Coster, Rushe, Holmes, Ventresque, and
  Dambre}]{de2023towards}
Mathieu De~Coster, Ellen Rushe, Ruth Holmes, Anthony Ventresque, and Joni
  Dambre. 2023.
\newblock Towards the extraction of robust sign embeddings for low resource
  sign language recognition.
\newblock \emph{arXiv preprint arXiv:2306.17558}.

\bibitem[{De~Sisto et~al.(2021)De~Sisto, Shterionov, Murtagh, Vermeerbergen,
  and Leeson}]{segmentation:de-sisto-etal-2021-defining}
Mirella De~Sisto, Dimitar Shterionov, Irene Murtagh, Myriam Vermeerbergen, and
  Lorraine Leeson. 2021.
\newblock \href {https://aclanthology.org/2021.mtsummit-at4ssl.11} {Defining
  meaningful units. challenges in sign segmentation and segment-meaning mapping
  (short paper)}.
\newblock In \emph{Proceedings of the 1st International Workshop on Automatic
  Translation for Signed and Spoken Languages (AT4SSL)}, pages 98--103,
  Virtual. Association for Machine Translation in the Americas.

\bibitem[{Devlin et~al.(2019)Devlin, Chang, Lee, and
  Toutanova}]{devlin-etal-2019-bert}
Jacob Devlin, Ming-Wei Chang, Kenton Lee, and Kristina Toutanova. 2019.
\newblock \href {https://doi.org/10.18653/v1/N19-1423} {{BERT}: Pre-training of
  deep bidirectional transformers for language understanding}.
\newblock In \emph{Proceedings of the 2019 Conference of the North {A}merican
  Chapter of the Association for Computational Linguistics: Human Language
  Technologies, Volume 1 (Long and Short Papers)}, pages 4171--4186,
  Minneapolis, Minnesota. Association for Computational Linguistics.

\bibitem[{Farag and Brock(2019)}]{segmentation:farag2019learning}
Iva Farag and Heike Brock. 2019.
\newblock Learning motion disfluencies for automatic sign language
  segmentation.
\newblock In \emph{ICASSP 2019-2019 IEEE International Conference on Acoustics,
  Speech and Signal Processing (ICASSP)}, pages 7360--7364. IEEE.

\bibitem[{Frost and Sutton(2022)}]{sw-hand-symbols}
Adam Frost and Valerie Sutton. 2022.
\newblock \emph{{SignWriting} Hand Symbols}.
\newblock SignWriting Press.

\bibitem[{Grishchenko and Bazarevsky(2020)}]{mediapipe2020holistic}
Ivan Grishchenko and Valentin Bazarevsky. 2020.
\newblock \href {https://google.github.io/mediapipe/solutions/holistic.html}
  {Mediapipe holistic}.

\bibitem[{Hanke et~al.(2012)Hanke, Matthes, Regen, and
  Worseck}]{Hanke2012WhereDA}
Thomas Hanke, Silke Matthes, Anja Regen, and Satu Worseck. 2012.
\newblock Where does a sign start and end? segmentation of continuous signing.
\newblock In \emph{Language Resources and Evaluation Conference}.

\bibitem[{Hanke et~al.(2020)Hanke, Schulder, Konrad, and
  Jahn}]{dataset:hanke-etal-2020-extending}
Thomas Hanke, Marc Schulder, Reiner Konrad, and Elena Jahn. 2020.
\newblock \href {https://www.aclweb.org/anthology/2020.signlang-1.12}
  {Extending the {P}ublic {DGS} {C}orpus in size and depth}.
\newblock In \emph{Proceedings of the LREC2020 9th Workshop on the
  Representation and Processing of Sign Languages: Sign Language Resources in
  the Service of the Language Community, Technological Challenges and
  Application Perspectives}, pages 75--82, Marseille, France. European Language
  Resources Association (ELRA).

\bibitem[{Hochreiter and Schmidhuber(1997)}]{hochreiter1997long}
Sepp Hochreiter and J{\"u}rgen Schmidhuber. 1997.
\newblock Long short-term memory.
\newblock \emph{Neural computation}, 9(8):1735--1780.

\bibitem[{Jiang et~al.(2023)Jiang, Soldati, Schamberg, Lameira, and
  Moran}]{jiang2023automatic}
Zifan Jiang, Adrian Soldati, Isaac Schamberg, Adriano~R Lameira, and Steven
  Moran. 2023.
\newblock Automatic sound event detection and classification of great ape calls
  using neural networks.
\newblock \emph{arXiv preprint arXiv:2301.02214}.

\bibitem[{Johnston(2008)}]{Johnston2008FromAT}
Trevor~Alexander Johnston. 2008.
\newblock From archive to corpus: transcription and annotation in the creation
  of signed language corpora.
\newblock In \emph{Pacific Asia Conference on Language, Information and
  Computation}.

\bibitem[{Liu et~al.(2021)Liu, Lin, Cao, Hu, Wei, Zhang, Lin, and
  Guo}]{liu2021swin}
Ze~Liu, Yutong Lin, Yue Cao, Han Hu, Yixuan Wei, Zheng Zhang, Stephen Lin, and
  Baining Guo. 2021.
\newblock Swin transformer: Hierarchical vision transformer using shifted
  windows.
\newblock In \emph{Proceedings of the IEEE/CVF international conference on
  computer vision}, pages 10012--10022.

\bibitem[{Mandel(1981)}]{Mandel1981PhonotacticsAM}
Mark~Alan Mandel. 1981.
\newblock \emph{Phonotactics and morphophonology in {A}merican {S}ign
  {L}anguage}.
\newblock University of California, Berkeley.

\bibitem[{Moryossef(2023)}]{moryossef2023addressing}
Amit Moryossef. 2023.
\newblock Addressing the blind spots in spoken language processing.
\newblock \emph{arXiv preprint arXiv:2309.06572}.

\bibitem[{Moryossef et~al.(2020)Moryossef, Tsochantaridis, Aharoni, Ebling, and
  Narayanan}]{detection:moryossef2020real}
Amit Moryossef, Ioannis Tsochantaridis, Roee Aharoni, Sarah Ebling, and Srini
  Narayanan. 2020.
\newblock Real-time sign-language detection using human pose estimation.
\newblock In \emph{Computer Vision--ECCV 2020 Workshops: Glasgow, UK, August
  23--28, 2020, Proceedings, Part II 16, SLRTP 2020: The Sign Language
  Recognition, Translation and Production Workshop}, pages 237--248.

\bibitem[{Neidle et~al.(2012)Neidle, Thangali, and
  Sclaroff}]{neidle2012challenges}
Carol Neidle, Ashwin Thangali, and Stan Sclaroff. 2012.
\newblock Challenges in development of the {A}merican {S}ign {L}anguage lexicon
  video dataset ({ASSLVD}) corpus.
\newblock In \emph{5th workshop on the representation and processing of sign
  languages: interactions between corpus and Lexicon, LREC}. Citeseer.

\bibitem[{Ormel and Crasborn(2012)}]{ormel2012prosodic}
Ellen Ormel and Onno Crasborn. 2012.
\newblock Prosodic correlates of sentences in signed languages: A literature
  review and suggestions for new types of studies.
\newblock \emph{Sign Language Studies}, 12(2):279--315.

\bibitem[{Pal et~al.(2023)Pal, Huber, Chaabani, Manzotti, and
  Koller}]{detection:pal2023importance}
Abhilash Pal, Stephan Huber, Cyrine Chaabani, Alessandro Manzotti, and Oscar
  Koller. 2023.
\newblock On the importance of signer overlap for sign language detection.
\newblock \emph{arXiv preprint arXiv:2303.10782}.

\bibitem[{Paszke et~al.(2019)Paszke, Gross, Massa, Lerer, Bradbury, Chanan,
  Killeen, Lin, Gimelshein, Antiga, Desmaison, Kopf, Yang, DeVito, Raison,
  Tejani, Chilamkurthy, Steiner, Fang, Bai, and Chintala}]{NEURIPS2019_9015}
Adam Paszke, Sam Gross, Francisco Massa, Adam Lerer, James Bradbury, Gregory
  Chanan, Trevor Killeen, Zeming Lin, Natalia Gimelshein, Luca Antiga, Alban
  Desmaison, Andreas Kopf, Edward Yang, Zachary DeVito, Martin Raison, Alykhan
  Tejani, Sasank Chilamkurthy, Benoit Steiner, Lu~Fang, Junjie Bai, and Soumith
  Chintala. 2019.
\newblock \href
  {http://papers.neurips.cc/paper/9015-pytorch-an-imperative-style-high-performance-deep-learning-library.pdf}
  {{PyTorch}: An imperative style, high-performance deep learning library}.
\newblock In \emph{Advances in Neural Information Processing Systems 32}, pages
  8024--8035. Curran Associates, Inc.

\bibitem[{Prillwitz et~al.(2008)Prillwitz, Hanke, K{\"o}nig, Konrad, Langer,
  and Schwarz}]{dataset:prillwitz2008dgs}
Siegmund Prillwitz, Thomas Hanke, Susanne K{\"o}nig, Reiner Konrad, Gabriele
  Langer, and Arvid Schwarz. 2008.
\newblock Dgs corpus project--development of a corpus based electronic
  dictionary german sign language/german.
\newblock In \emph{sign-lang at LREC 2008}, pages 159--164. European Language
  Resources Association (ELRA).

\bibitem[{Prillwitz and Zienert(1990)}]{writing:prillwitz1990hamburg}
Siegmund Prillwitz and Heiko Zienert. 1990.
\newblock Hamburg notation system for sign language: Development of a sign
  writing with computer application.
\newblock In \emph{Current trends in European Sign Language Research.
  Proceedings of the 3rd European Congress on Sign Language Research}, pages
  355--379.

\bibitem[{Que et~al.(2022)Que, Nakahara, Fan, Li, Meng, Tsoi, Niu, Nurvitadhi,
  and Luk}]{Que2022RemarnAR}
Zhiqiang Que, Hiroki Nakahara, Hongxiang Fan, He~Li, Jiuxi Meng, Kuen~Hung
  Tsoi, Xinyu Niu, Eriko Nurvitadhi, and Wayne W.~C. Luk. 2022.
\newblock Remarn: A reconfigurable multi-threaded multi-core accelerator for
  recurrent neural networks.
\newblock \emph{ACM Transactions on Reconfigurable Technology and Systems},
  16:1 -- 26.

\bibitem[{Que et~al.(2020)Que, Nakahara, Nurvitadhi, Fan, Zeng, Meng, Niu, and
  Luk}]{Que2020OptimizingRR}
Zhiqiang Que, Hiroki Nakahara, Eriko Nurvitadhi, Hongxiang Fan, Chenglong Zeng,
  Jiuxi Meng, Xinyu Niu, and Wayne W.~C. Luk. 2020.
\newblock Optimizing reconfigurable recurrent neural networks.
\newblock \emph{2020 IEEE 28th Annual International Symposium on
  Field-Programmable Custom Computing Machines (FCCM)}, pages 10--18.

\bibitem[{Que et~al.(2021)Que, Wang, Marikar, Moreno, Ngadiuba, Javed,
  Borzyszkowski, Aarrestad, Loncar, Summers, Pierini, Cheung, and
  Luk}]{Que2021AcceleratingRN}
Zhiqiang Que, Erwei Wang, Umar Marikar, Eric Moreno, Jennifer Ngadiuba, Hamza
  Javed, Bartłomiej Borzyszkowski, Thea~Klaeboe Aarrestad, Vladimir Loncar,
  Sioni~Paris Summers, Maurizio Pierini, Peter Y.~K. Cheung, and Wayne W.~C.
  Luk. 2021.
\newblock Accelerating recurrent neural networks for gravitational wave
  experiments.
\newblock \emph{2021 IEEE 32nd International Conference on Application-specific
  Systems, Architectures and Processors (ASAP)}, pages 117--124.

\bibitem[{Ram{\i}rez et~al.(2004)Ram{\i}rez, Segura, Ben{\i}tez, De~La~Torre,
  and Rubio}]{ramirez2004efficient}
Javier Ram{\i}rez, Jos{\'e}~C Segura, Carmen Ben{\i}tez, Angel De~La~Torre, and
  Antonio Rubio. 2004.
\newblock Efficient voice activity detection algorithms using long-term speech
  information.
\newblock \emph{Speech communication}, 42(3-4):271--287.

\bibitem[{Ramshaw and Marcus(1995)}]{ramshaw-marcus-1995-text}
Lance Ramshaw and Mitch Marcus. 1995.
\newblock \href {https://aclanthology.org/W95-0107} {Text chunking using
  transformation-based learning}.
\newblock In \emph{Third Workshop on Very Large Corpora}.

\bibitem[{Redmon et~al.(2015)Redmon, Divvala, Girshick, and
  Farhadi}]{Redmon2015YouOL}
Joseph Redmon, Santosh~Kumar Divvala, Ross~B. Girshick, and Ali Farhadi. 2015.
\newblock You only look once: Unified, real-time object detection.
\newblock \emph{2016 IEEE Conference on Computer Vision and Pattern Recognition
  (CVPR)}, pages 779--788.

\bibitem[{Renz et~al.(2021{\natexlab{a}})Renz, Stache, Albanie, and
  Varol}]{segmentation:renz2021signa}
Katrin Renz, Nicolaj~C Stache, Samuel Albanie, and G{\"u}l Varol.
  2021{\natexlab{a}}.
\newblock Sign language segmentation with temporal convolutional networks.
\newblock In \emph{ICASSP 2021-2021 IEEE International Conference on Acoustics,
  Speech and Signal Processing (ICASSP)}, pages 2135--2139. IEEE.

\bibitem[{Renz et~al.(2021{\natexlab{b}})Renz, Stache, Fox, Varol, and
  Albanie}]{segmentation:renz2021signb}
Katrin Renz, Nicolaj~C Stache, Neil Fox, Gul Varol, and Samuel Albanie.
  2021{\natexlab{b}}.
\newblock Sign segmentation with changepoint-modulated pseudo-labelling.
\newblock In \emph{Proceedings of the IEEE/CVF Conference on Computer Vision
  and Pattern Recognition}, pages 3403--3412.

\bibitem[{Sandler(2010)}]{sandler2010prosody}
Wendy Sandler. 2010.
\newblock Prosody and syntax in sign languages.
\newblock \emph{Transactions of the philological society}, 108(3):298--328.

\bibitem[{Sandler and Lillo-Martin(2006)}]{sandler2006sign}
Wendy Sandler and Diane Lillo-Martin. 2006.
\newblock \emph{Sign language and linguistic universals}.
\newblock Cambridge University Press.

\bibitem[{Sandler et~al.(2008)Sandler, Macneilage, Davis, Zajdo, York, and
  Francis}]{Sandler2008TheSI}
Wendy Sandler, Peter Macneilage, Barbara Davis, Kristine Zajdo, New York, and
  Taylor Francis. 2008.
\newblock \href
  {http://sandlersignlab.haifa.ac.il/pdf/The_syllable_in_sign_language.pdf}
  {The syllable in sign language: Considering the other natural language
  modality}.

\bibitem[{Santemiz et~al.(2009)Santemiz, Aran, Saraclar, and
  Akarun}]{segmentation:santemiz2009automatic}
Pinar Santemiz, Oya Aran, Murat Saraclar, and Lale Akarun. 2009.
\newblock Automatic sign segmentation from continuous signing via multiple
  sequence alignment.
\newblock In \emph{2009 IEEE 12th International Conference on Computer Vision
  Workshops, ICCV Workshops}, pages 2001--2008. IEEE.

\bibitem[{Schulder and Hanke(2019)}]{schulderopenpose}
Marc Schulder and Thomas Hanke. 2019.
\newblock \href {https://doi.org/10.25592/uhhfdm.842} {{OpenPose} in the
  {Public} {DGS} {Corpus}}.
\newblock Project Note AP06-2019-01, DGS-Korpus project, IDGS, Hamburg
  University, Hamburg, Germany.

\bibitem[{Sehyr et~al.(2021)Sehyr, Caselli, Cohen-Goldberg, and
  Emmorey}]{sehyr2021asl}
Zed~Sevcikova Sehyr, Naomi Caselli, Ariel~M Cohen-Goldberg, and Karen Emmorey.
  2021.
\newblock The {ASL-LEX} 2.0 project: A database of lexical and phonological
  properties for 2,723 signs in american sign language.
\newblock \emph{The Journal of Deaf Studies and Deaf Education},
  26(2):263--277.

\bibitem[{Simonyan and Zisserman(2015)}]{simonyan2015very}
Karen Simonyan and Andrew Zisserman. 2015.
\newblock Very deep convolutional networks for large-scale image recognition.
\newblock \emph{CoRR}.

\bibitem[{Sohn et~al.(1999)Sohn, Kim, and Sung}]{sohn1999statistical}
Jongseo Sohn, Nam~Soo Kim, and Wonyong Sung. 1999.
\newblock A statistical model-based voice activity detection.
\newblock \emph{IEEE signal processing letters}, 6(1):1--3.

\bibitem[{Stokoe~Jr(1960)}]{writing:stokoe1960sign}
William~C Stokoe~Jr. 1960.
\newblock \href {https://doi.org/10.1093/deafed/eni001} {{Sign Language
  Structure: An Outline of the Visual Communication Systems of the American
  Deaf}}.
\newblock \emph{The Journal of Deaf Studies and Deaf Education}, 10(1):3--37.

\bibitem[{Sutton(1990)}]{writing:sutton1990lessons}
Valerie Sutton. 1990.
\newblock \emph{Lessons in sign writing}.
\newblock SignWriting.

\bibitem[{Tang et~al.(2019)Tang, Ding, Rao, Zheng, Zhang, Zhao, Lu, and
  Zhou}]{tang2019coin}
Yansong Tang, Dajun Ding, Yongming Rao, Yu~Zheng, Danyang Zhang, Lili Zhao,
  Jiwen Lu, and Jie Zhou. 2019.
\newblock {COIN}: A large-scale dataset for comprehensive instructional video
  analysis.
\newblock In \emph{IEEE Conference on Computer Vision and Pattern Recognition
  (CVPR)}.

\bibitem[{Tsiamas et~al.(2022)Tsiamas, Gállego, Fonollosa, and
  Costa-jussà}]{tsiamas22_interspeech}
Ioannis Tsiamas, Gerard~I. Gállego, José A.~R. Fonollosa, and Marta~R.
  Costa-jussà. 2022.
\newblock \href {https://doi.org/10.21437/Interspeech.2022-59} {{SHAS}:
  Approaching optimal segmentation for end-to-end speech translation}.
\newblock In \emph{Proc. Interspeech 2022}, pages 106--110.

\bibitem[{Varol et~al.(2022)Varol, Momeni, Albanie, Afouras, and
  Zisserman}]{Varol2022ScalingUS}
G{\"u}l Varol, Liliane Momeni, Samuel Albanie, Triantafyllos Afouras, and
  Andrew Zisserman. 2022.
\newblock Scaling up sign spotting through sign language dictionaries.
\newblock \emph{International Journal of Computer Vision}, 130:1416 -- 1439.

\bibitem[{Vaswani et~al.(2017)Vaswani, Shazeer, Parmar, Uszkoreit, Jones,
  Gomez, Kaiser, and Polosukhin}]{vaswani2017attention}
Ashish Vaswani, Noam Shazeer, Niki Parmar, Jakob Uszkoreit, Llion Jones,
  Aidan~N Gomez, {\L}ukasz Kaiser, and Illia Polosukhin. 2017.
\newblock Attention is all you need.
\newblock \emph{Advances in neural information processing systems}, 30.

\bibitem[{Venkatesh et~al.(2022)Venkatesh, Moffat, and
  Miranda}]{venkatesh2022you}
Satvik Venkatesh, David Moffat, and Eduardo~Reck Miranda. 2022.
\newblock You only hear once: a {YOLO}-like algorithm for audio segmentation
  and sound event detection.
\newblock \emph{Applied Sciences}, 12(7):3293.

\bibitem[{Wong et~al.(2022)Wong, Camgoz, and Bowden}]{Wong2022HierarchicalIF}
Ryan Wong, Necati~Cihan Camgoz, and R.~Bowden. 2022.
\newblock Hierarchical {I3D} for sign spotting.
\newblock In \emph{ECCV Workshops}.

\bibitem[{Yan et~al.(2018)Yan, Xiong, and Lin}]{yan2018spatial}
Sijie Yan, Yuanjun Xiong, and Dahua Lin. 2018.
\newblock Spatial temporal graph convolutional networks for skeleton-based
  action recognition.
\newblock In \emph{Proceedings of the AAAI conference on artificial
  intelligence}, volume~32.

\bibitem[{Yu et~al.(2017)Yu, Yin, and Zhu}]{Yu2017SpatioTemporalGC}
Ting Yu, Haoteng Yin, and Zhanxing Zhu. 2017.
\newblock Spatio-temporal graph convolutional networks: A deep learning
  framework for traffic forecasting.
\newblock In \emph{International Joint Conference on Artificial Intelligence}.

\bibitem[{Zhang et~al.(2023)Zhang, M{\"u}ller, and Sennrich}]{zhang2023sltunet}
Biao Zhang, Mathias M{\"u}ller, and Rico Sennrich. 2023.
\newblock \href {https://openreview.net/forum?id=EBS4C77p_5S} {{SLTUNET}: A
  simple unified model for sign language translation}.
\newblock In \emph{The Eleventh International Conference on Learning
  Representations}, Kigali, Rwanda.

\end{thebibliography}
\bibliographystyle{acl_natbib}

\onecolumn
\newpage

\appendix

\section{Extended Experimental Results}
\label{appendix:extended-results}

We conducted some preliminary experiments (starting with \textit{P0}) on training a sign language segmentation model to gain insights into hyperparameters and feature choices. The results are shown in Table \ref{tab:preliminary}\footnote{Note that due to an implementation issue on edge cases (which we fixed later), the \emph{IoU} and \emph{\%} values in Table \ref{tab:preliminary} are lower than the ones in Table \ref{tab:results} and Table \ref{tab:results-full} thus not comparable across tables. The comparison inside of Table \ref{tab:preliminary} between different experiments remains meaningful. In addition, the results in Table \ref{tab:preliminary} are based on only one run instead of three random runs.}. We found in \textit{P1.3.2} the optimal hyperparameters and repeated them with different feature choices.

\begin{table*}[htbp]
\small
\centering
\begin{tabular}{lllccc|ccc}
\toprule
& & & \multicolumn{3}{c}{\textbf{Sign}} & \multicolumn{3}{c}{\textbf{Phrase}}\\
\cmidrule(lr){4-6} \cmidrule(lr){7-9}
\multicolumn{3}{l}{\textbf{Experiment}} & \textbf{F1} & \textbf{IoU} & \textbf{\%} & \textbf{F1} & \textbf{IoU} & \textbf{\%}\\
\midrule
\textbf{P0} & \textbf{Moryossef et al. (2020)} & \textbf{test} &--- & $0.4$ & $1.45$ & --- & $0.65$ & $0.82$\\
& & \textbf{dev} &--- & $0.35$ & $1.36$ & --- & $0.6$ & $0.77$\\
\textbf{P0.1} & \textbf{P0 + Holistic 25fps} & \textbf{test} &--- & $0.39$ & $0.86$ & --- & $0.64$ & $0.5$\\
& & \textbf{dev} &--- & $0.32$ & $0.81$ & --- & $0.58$ & $0.52$\\
\midrule
\textbf{P1} & \textbf{P1 baseline} & \textbf{test} &$0.55$ & $0.49$ & $0.83$ & $0.6$ & $0.67$ & $2.63$\\
& & \textbf{dev} &$0.56$ & $0.43$ & $0.75$ & $0.58$ & $0.62$ & $2.61$\\
\textbf{P1.1} & \textbf{P1 - encoder\_bidirectional} & \textbf{test} &$0.48$ & $0.45$ & $0.68$ & $0.5$ & $0.64$ & $2.68$\\
& & \textbf{dev} &$0.46$ & $0.41$ & $0.64$ & $0.51$ & $0.61$ & $2.56$\\
\textbf{P1.2.1} & \textbf{P1 + hidden\_dim=512} & \textbf{test} &$0.47$ & $0.42$ & $0.44$ & $0.52$ & $0.63$ & $1.7$\\
& & \textbf{dev} &$0.46$ & $0.4$ & $0.43$ & $0.52$ & $0.61$ & $1.69$\\
\textbf{P1.2.2} & \textbf{P1 + hidden\_dim=1024} & \textbf{test} &$0.48$ & $0.45$ & $0.42$ & $0.58$ & $0.65$ & $1.53$\\
& & \textbf{dev} &$0.46$ & $0.41$ & $0.36$ & $0.53$ & $0.61$ & $1.49$\\
\textbf{P1.3.1} & \textbf{P1 + encoder\_depth=2} & \textbf{test} &$0.55$ & $0.48$ & $0.76$ & $0.58$ & $0.67$ & $2.56$\\
& & \textbf{dev} &$0.56$ & $0.43$ & $0.69$ & $0.58$ & $0.62$ & $2.52$\\
\textbf{P1.3.2} & \textbf{P1 + encoder\_depth=4} & \textbf{test} &$0.63$ & $0.51$ & $0.91$ & $0.66$ & $0.67$ & $1.41$\\
& & \textbf{dev} &$0.61$ & $0.47$ & $0.84$ & $0.64$ & $0.6$ & $1.39$\\
\textbf{P1.4.1} & \textbf{P1 + hidden\_dim=128 + encoder\_depth=2} & \textbf{test} &$0.58$ & $0.48$ & $0.8$ & $0.6$ & $0.67$ & $2.0$\\
& & \textbf{dev} &$0.55$ & $0.43$ & $0.75$ & $0.54$ & $0.62$ & $2.03$\\
\textbf{P1.4.2} & \textbf{P1 + hidden\_dim=128 + encoder\_depth=4} & \textbf{test} &$0.62$ & $0.51$ & $0.91$ & $0.64$ & $0.68$ & $2.43$\\
& & \textbf{dev} &$0.6$ & $0.47$ & $0.83$ & $0.6$ & $0.62$ & $2.57$\\
\textbf{P1.4.3} & \textbf{P1 + hidden\_dim=128 + encoder\_depth=8} & \textbf{test} &$0.59$ & $0.52$ & $0.91$ & $0.63$ & $0.68$ & $3.04$\\
& & \textbf{dev} &$0.6$ & $0.47$ & $0.84$ & $0.6$ & $0.62$ & $3.02$\\
\textbf{P1.5.1} & \textbf{P1 + hidden\_dim=64 + encoder\_depth=4} & \textbf{test} &$0.57$ & $0.5$ & $0.8$ & $0.6$ & $0.68$ & $2.41$\\
& & \textbf{dev} &$0.58$ & $0.45$ & $0.75$ & $0.59$ & $0.62$ & $2.39$\\
\textbf{P1.5.2} & \textbf{P1 + hidden\_dim=64 + encoder\_depth=8} & \textbf{test} &$0.62$ & $0.51$ & $0.85$ & $0.64$ & $0.68$ & $2.53$\\
& & \textbf{dev} &$0.6$ & $0.46$ & $0.79$ & $0.6$ & $0.62$ & $2.53$\\
\midrule
\textbf{P2} & \textbf{P1 + optical\_flow} & \textbf{test} &$0.58$ & $0.5$ & $0.95$ & $0.63$ & $0.68$ & $3.17$\\
& & \textbf{dev} &$0.59$ & $0.45$ & $0.84$ & $0.59$ & $0.61$ & $3.08$\\
\textbf{P2.1} & \textbf{P1.3.2 + optical\_flow} & \textbf{test} &$0.63$ & $0.51$ & $0.92$ & $0.66$ & $0.67$ & $1.51$\\
& & \textbf{dev} &$0.62$ & $0.46$ & $0.81$ & $0.62$ & $0.6$ & $1.53$\\
\midrule
\textbf{P3} & \textbf{P1 + hand\_normalization} & \textbf{test} &$0.55$ & $0.48$ & $0.77$ & $0.58$ & $0.67$ & $2.79$\\
& & \textbf{dev} &$0.55$ & $0.42$ & $0.71$ & $0.57$ & $0.62$ & $2.73$\\
\textbf{P3.1} & \textbf{P1.3.2 + hand\_normalization} & \textbf{test} &$0.63$ & $0.51$ & $0.91$ & $0.66$ & $0.66$ & $1.43$\\
& & \textbf{dev} &$0.61$ & $0.46$ & $0.82$ & $0.64$ & $0.61$ & $1.46$\\
\midrule
\textbf{P4} & \textbf{P2.1 + P3.1} & \textbf{test} &$0.56$ & $0.51$ & $0.92$ & $0.61$ & $0.66$ & $1.45$\\
& & \textbf{dev} &$0.61$ & $0.46$ & $0.81$ & $0.63$ & $0.6$ & $1.41$\\
\textbf{P4.1} & \textbf{P4 + encoder\_depth=8} & \textbf{test} &$0.6$ & $0.51$ & $0.95$ & $0.62$ & $0.67$ & $1.08$\\
& & \textbf{dev} &$0.61$ & $0.47$ & $0.86$ & $0.62$ & $0.6$ & $1.12$\\
\midrule
\textbf{P5} & \textbf{P1.3.2 + reduced\_face} & \textbf{test} &$0.63$ & $0.51$ & $0.94$ & $0.64$ & $0.66$ & $1.16$\\
& & \textbf{dev} &$0.61$ & $0.47$ & $0.86$ & $0.64$ & $0.58$ & $1.14$\\
\textbf{P5.1} & \textbf{P1.3.2 + full\_face} & \textbf{test} &$0.54$ & $0.49$ & $0.8$ & $0.6$ & $0.68$ & $2.29$\\
& & \textbf{dev} &$0.57$ & $0.45$ & $0.7$ & $0.59$ & $0.62$ & $2.29$\\

\bottomrule
\end{tabular}
\caption{Results of the preliminary experiments.}
\label{tab:preliminary}
\end{table*}

\clearpage
We selected some promising models from our preliminary experiments and reran them three times using different random seeds to make the final conclusion reliable and robust. Table \ref{tab:results-full} includes the standard deviation and the validation results (where we performed the model selection) for readers to scrutinize.

\begin{table*}[htbp]
% \small
\centering
\resizebox{\textwidth}{!}{%
\begin{tabular}{lllccc|ccc|cc}
\toprule
& & & \multicolumn{3}{c}{\textbf{Sign}} & \multicolumn{3}{c}{\textbf{Phrase}} & \multicolumn{2}{c}{\textbf{Efficiency}} \\
\cmidrule(lr){4-6} \cmidrule(lr){7-9} \cmidrule(lr){10-11}
\multicolumn{3}{l}{\textbf{Experiment}} & \textbf{F1} & \textbf{IoU} & \textbf{\%} & \textbf{F1} & \textbf{IoU} & \textbf{\%} & \textbf{\#Params} & \textbf{Time} \\
\midrule

\textbf{E0} & \textbf{\citet{detection:moryossef2020real}} & \textbf{test} &--- & $0.46\pm0.03$ & $1.09\pm0.41$ & --- & $0.70\pm0.01$ & $1.00\pm0.06$ & 102K & 0:50:17\\
& & \textbf{dev} &--- & $0.42\pm0.05$ & $1.21\pm0.59$ & --- & $0.61\pm0.06$ & $2.47\pm0.85$ & 102K & 0:50:17\\
\midrule
\textbf{E1} & \textbf{Baseline} & \textbf{test} &$0.56\pm0.03$ & $0.66\pm0.01$ & $0.91\pm0.05$ & $0.59\pm0.02$ & $0.80\pm0.03$ & $2.50\pm0.13$ & 454K & 1:01:50\\
& & \textbf{dev} &$0.55\pm0.01$ & $0.59\pm0.00$ & $1.12\pm0.11$ & $0.56\pm0.02$ & $0.75\pm0.05$ & $2.94\pm0.08$ & 454K & 1:01:50\\
\textbf{E2} & \textbf{E1 + Face} & \textbf{test} &$0.53\pm0.05$ & $0.58\pm0.07$ & $0.64\pm0.30$ & $0.57\pm0.02$ & $0.76\pm0.03$ & $1.87\pm0.83$ & 552K & 1:50:31\\
& & \textbf{dev} &$0.50\pm0.07$ & $0.53\pm0.11$ & $0.90\pm0.19$ & $0.53\pm0.05$ & $0.71\pm0.07$ & $2.43\pm1.02$ & 552K & 1:50:31\\
\textbf{E3} & \textbf{E1 + Optical Flow} & \textbf{test} &$0.58\pm0.01$ & $0.62\pm0.00$ & $1.12\pm0.05$ & $0.60\pm0.03$ & $0.82\pm0.03$ & $3.19\pm0.11$ & 473K & 1:20:17\\
& & \textbf{dev} &$0.58\pm0.00$ & $0.62\pm0.00$ & $1.50\pm0.19$ & $0.59\pm0.01$ & $0.79\pm0.00$ & $3.94\pm0.14$ & 473K & 1:20:17\\
\textbf{E4} & \textbf{E3 + Hand Norm} & \textbf{test} &$0.56\pm0.02$ & $0.61\pm0.00$ & $1.07\pm0.05$ & $0.60\pm0.00$ & $0.80\pm0.00$ & $3.24\pm0.17$ & 516K & 1:30:59\\
& & \textbf{dev} &$0.57\pm0.01$ & $0.61\pm0.01$ & $1.50\pm0.07$ & $0.58\pm0.00$ & $0.79\pm0.00$ & $4.04\pm0.31$ & 516K & 1:30:59\\
\midrule
\textbf{E1s} & \textbf{E1 + Depth=4} & \textbf{test} &$0.63\pm0.01$ & $0.69\pm0.00$ & $1.11\pm0.01$ & $0.65\pm0.02$ & $0.82\pm0.04$ & $1.63\pm0.10$ & 1.6M & 4:08:48\\
& & \textbf{dev} &$0.61\pm0.00$ & $0.63\pm0.00$ & $1.27\pm0.01$ & $0.63\pm0.01$ & $0.77\pm0.01$ & $2.17\pm0.18$ & 1.6M & 4:08:48\\
\textbf{E2s} & \textbf{E2 + Depth=4} & \textbf{test} &$0.62\pm0.02$ & $0.69\pm0.00$ & $1.07\pm0.03$ & $0.63\pm0.01$ & $0.84\pm0.03$ & $2.68\pm0.53$ & 1.7M & 3:14:03\\
& & \textbf{dev} &$0.60\pm0.01$ & $0.63\pm0.01$ & $1.20\pm0.12$ & $0.59\pm0.02$ & $0.76\pm0.05$ & $3.30\pm0.62$ & 1.7M & 3:14:03\\
\textbf{E3s} & \textbf{E3 + Depth=4} & \textbf{test} &$0.60\pm0.01$ & $0.63\pm0.00$ & $1.13\pm0.01$ & $0.64\pm0.03$ & $0.80\pm0.03$ & $1.53\pm0.18$ & 1.7M & 4:08:30\\
& & \textbf{dev} &$0.62\pm0.00$ & $0.63\pm0.00$ & $1.63\pm0.05$ & $0.63\pm0.00$ & $0.76\pm0.00$ & $2.14\pm0.09$ & 1.7M & 4:08:30\\
\textbf{E4s} & \textbf{E4 + Depth=4} & \textbf{test} &$0.59\pm0.00$ & $0.63\pm0.00$ & $1.13\pm0.03$ & $0.62\pm0.00$ & $0.79\pm0.00$ & $1.43\pm0.10$ & 1.7M & 4:35:29\\
& & \textbf{dev} &$0.61\pm0.00$ & $0.63\pm0.00$ & $1.56\pm0.04$ & $0.63\pm0.00$ & $0.77\pm0.01$ & $1.89\pm0.07$ & 1.7M & 4:35:29\\
\midrule
\textbf{E4ba} & \textbf{E4s + Autoregressive} & \textbf{test} &$0.45\pm0.03$ & $0.47\pm0.05$ & $0.88\pm0.08$ & $0.52\pm0.02$ & $0.63\pm0.10$ & $2.72\pm1.33$ & 1.3M & 2 days, 21:28:42\\
& & \textbf{dev} &$0.40\pm0.01$ & $0.40\pm0.01$ & $2.02\pm0.73$ & $0.47\pm0.00$ & $0.57\pm0.04$ & $4.26\pm1.26$ & 1.3M & 2 days, 21:28:42\\

\bottomrule
\end{tabular}
}
\caption{Mean evaluation metrics for our main experiments. A complete version of Table \ref{tab:results}.}
\label{tab:results-full}
\end{table*}

\clearpage
\section{Greedy Decoding Algorithm}\label{appendix:greedy-decoding}

We provide our exact decoding algorithm in Algorithm \ref{algo:probs_to_segments}.
We opt to employ adjustable thresholds rather than \emph{argmax} prediction, as our empirical findings demonstrate superior performance with this approach (\S\ref{sec:tuning-decoding}).

\begin{algorithm}
\caption{Probabilities to Segments Conversion.}
\begin{algorithmic}[1]
\label{algo:probs_to_segments}
\REQUIRE $probs$, a list of probabilities from 0 to 100
\STATE $threshold_b \gets 50.0$
\STATE $threshold_o \gets 50.0$
\STATE
\STATE $start \gets None$
\STATE $did\_pass\_start \gets False$
\STATE
\FOR{$i = 0$ \TO $len(probs)$}
    \STATE $b, i, o \gets probs[i]$
    \STATE
    \IF{$start = None$}
        \IF{$b > threshold_b$}
            \STATE $start \gets i$
        \ENDIF
    \ELSE
        \IF{$did\_pass\_start$}
            \IF{$b > threshold_b$ \OR $o > threshold_o$}
                \STATE \textbf{yield} $(start, i - 1))$

                \STATE $start \gets None$
                \STATE $did\_pass\_start \gets False$
            \ENDIF
        \ELSE
            \IF{$b < threshold_b$}
                \STATE $did\_pass\_start \gets True$
            \ENDIF
        \ENDIF
    \ENDIF
\ENDFOR
\STATE
\IF{$start \neq None$}
    \STATE \textbf{yield} $(start,  len(probs)))$
\ENDIF
\end{algorithmic}
\end{algorithm}

\newpage

\section{Pose Based Hand Shape Analysis}\label{appendix:hand-analysis}

\subsection{Introduction to Hand Shapes in Sign Language}

The most prominent feature of signed languages is their use of the hands. In fact, the hands play an integral role in the phonetics of signs, and a slight variation in hand shape can convey differences in meaning \cite{writing:stokoe1960sign}.
In sign languages such as American Sign Language (ASL) and British Sign Language (BSL), different hand shapes contribute to the vocabulary of the language, similar to how different sounds contribute to the vocabulary of spoken languages. ASL is estimated to use between 30 to 80 hand shapes\footnote{\url{https://aslfont.github.io/Symbol-Font-For-ASL/asl/handshapes.html}}, while BSL is limited to approximately 40 hand shapes\footnote{\url{https://bsl.surrey.ac.uk/principles/i-hand-shapes}}.
SignWriting \cite{writing:sutton1990lessons}, a system of notation used for sign languages, specifies a superset of 261 distinct hand shapes \cite{sw-hand-symbols}. Each sign language uses a subset of these hand shapes.

Despite the fundamental role of hand shapes in sign languages, accurately recognizing and classifying them is a challenging task.
In this section, we explore rule-based hand shape analysis in sign languages using 3D hand normalization. By performing 3D hand normalization, we can transform any given hand shape to a fixed orientation, making it easier for a model to extract the hand shape, and hence improving the recognition and classification of hand shapes in sign languages.

\subsection{Characteristics of the Human Hand}

The human hand consists of 27 bones and can be divided into three main sections: the wrist (carpals), the palm (metacarpals), and the fingers (phalanges). Each finger consists of three bones, except for the thumb, which has two. The bones are connected by joints, which allow for the complex movements and shapes that the hand can form.

\begin{figure}[h!]
\centering
\includegraphics[width=0.75\textwidth]{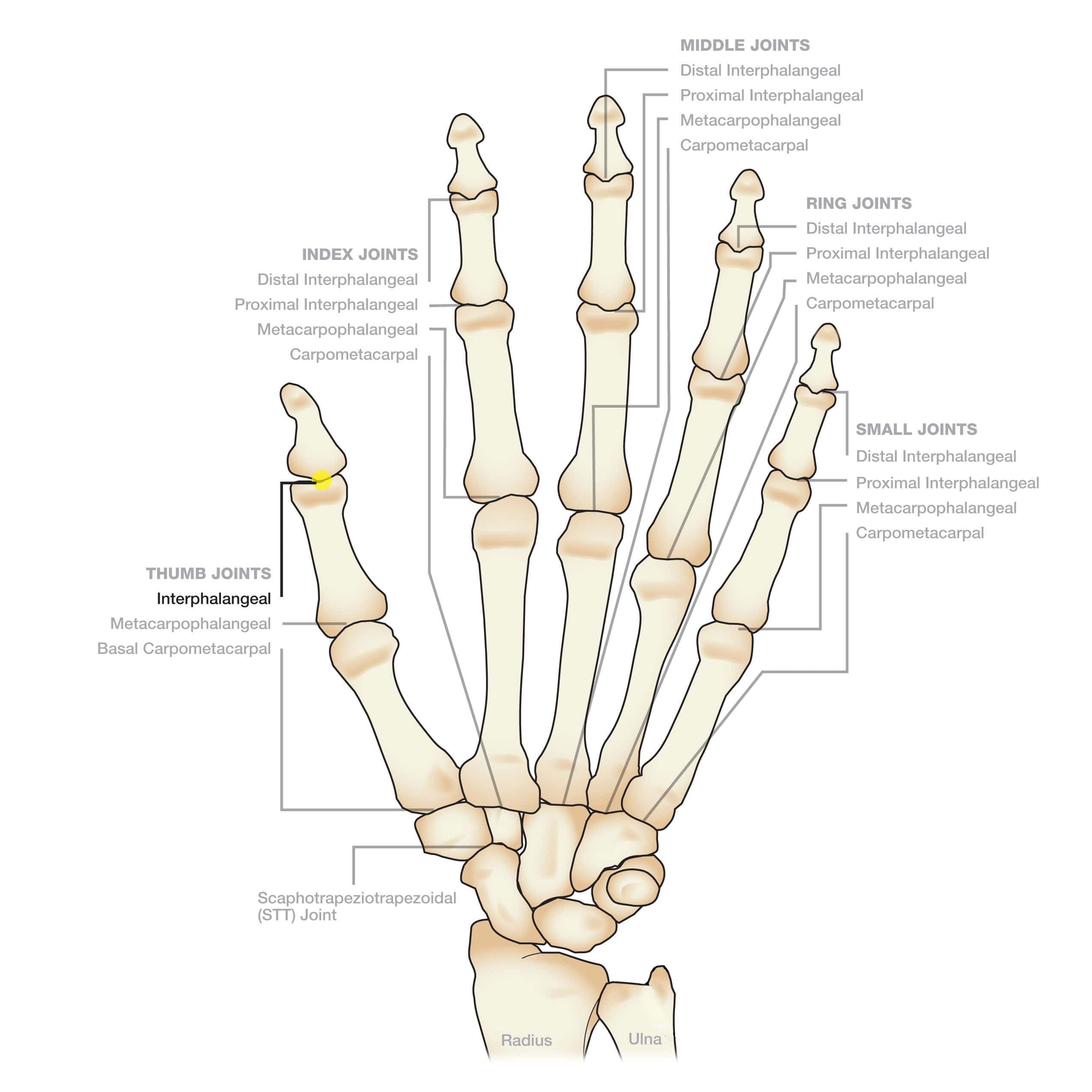}
\caption{Anatomy of a human hand. \copyright American Society for Surgery of the Hand}
\label{fig:hand_anatomy}
\end{figure}

Understanding the different characteristics of hands and their implications in signed languages is crucial for the extraction and classification of hand shapes. These characteristics are based on the SignWriting definitions of the five major axes of hand variation: handedness, plane, rotation, view, and shape. 

\paragraph{Handedness} is the distinction between the right and left hands. Signed languages make a distinction between the dominant hand and the non-dominant hand. For right-handed individuals, the right hand is considered dominant, and vice-versa. The dominant hand is used for fingerspelling and all one-handed signs, while the non-dominant hand is used for support and two-handed signs. Using 3D pose estimation, the handedness analysis is trivial, as the pose estimation platform predicts which hand is which.

\paragraph{Plane} refers to whether the hand is parallel to the wall or the floor. The variation in the plane can, but does not have to, create a distinction between two signs. For example, in ASL the signs for ``date'' and ``dessert'' exhibit the same hand shape, view, rotation, and movement, but differ by plane. The plane of a hand can be estimated by comparing the positions of the wrist and middle finger metacarpal bone ($M\_MCP$).

\begin{algorithm}
\caption{Hand Plane Estimation}
\begin{algorithmic}[1]  % The number tells where the line numbering should start
    \STATE $y \gets |M\_MCP.y - WRIST.y| \times 1.5$ // add bias to y
    \STATE $z \gets |M\_MCP.z - WRIST.z|$
    \RETURN $y > z$ ? `wall' : `floor'
\end{algorithmic}
\end{algorithm}

\paragraph{Rotation} refers to the angle of the hand in relation to the body. SignWriting groups the hand rotation into eight equal categories, each spanning 45 degrees. The rotation of a hand can be calculated by finding the angle of the line created by the wrist and the middle finger metacarpal bone.

\paragraph{View} refers to the side of the hand as observed by the signer, and is grouped into four categories: front, back, sideways, and other-sideways. The view of a hand can be estimated by analyzing the normal of the plane created by the palm of the hand (between the wrist, index finger metacarpal bone, and pinky metacarpal bone).

\begin{algorithm}
\caption{Hand View Estimation}
\begin{algorithmic}[1]
\STATE $\text{normal} \gets \text{math.normal}(\text{WRIST}, \text{I\_MCP}, \text{P\_MCP})$
\STATE $\text{plane} \gets \text{get\_plane}(\text{WRIST}, \text{M\_MCP})$
\IF{plane = `wall'}
    \STATE $angle \gets \angle(normal.z, normal.x)$
    \RETURN $angle > 210$ ? `front' : ($angle > 150$ ? `sideways' : `back')
\ELSE
    \STATE $angle \gets \angle(normal.y, normal.x)$
    \RETURN $angle > 0$ ? `front' : ($angle > -60$ ? `sideways' : `back')
\ENDIF
\end{algorithmic}
\end{algorithm}

\paragraph{Shape} refers to the configuration of the fingers and thumb. This characteristic of the hand is the most complex to analyze due to the vast array of possible shapes the human hand can form. The shape of a hand is determined by the state of each finger and thumb, specifically whether they are straight, curved, or bent, and their position relative to each other. 
Shape analysis can be accomplished by examining the bend and rotation of each finger joint. More advanced models may also take into consideration the spread between the fingers and other nuanced characteristics. 3D pose estimation can be used to extract these features for a machine learning model, which can then classify the hand shape.

\subsection{3D Hand Normalization}

3D hand normalization is an attempt at standardizing the orientation and position of the hand, thereby enabling models to effectively classify various hand shapes. The normalization process involves several key steps, as illustrated below:

\begin{enumerate}
\item \textbf{Pose Estimation} Initially, the 3D pose of the hand is estimated from the hand image crop (Figure \ref{fig:original_hands_poses}).

\begin{figure}[h!]
\centering
\includegraphics[width=\textwidth]{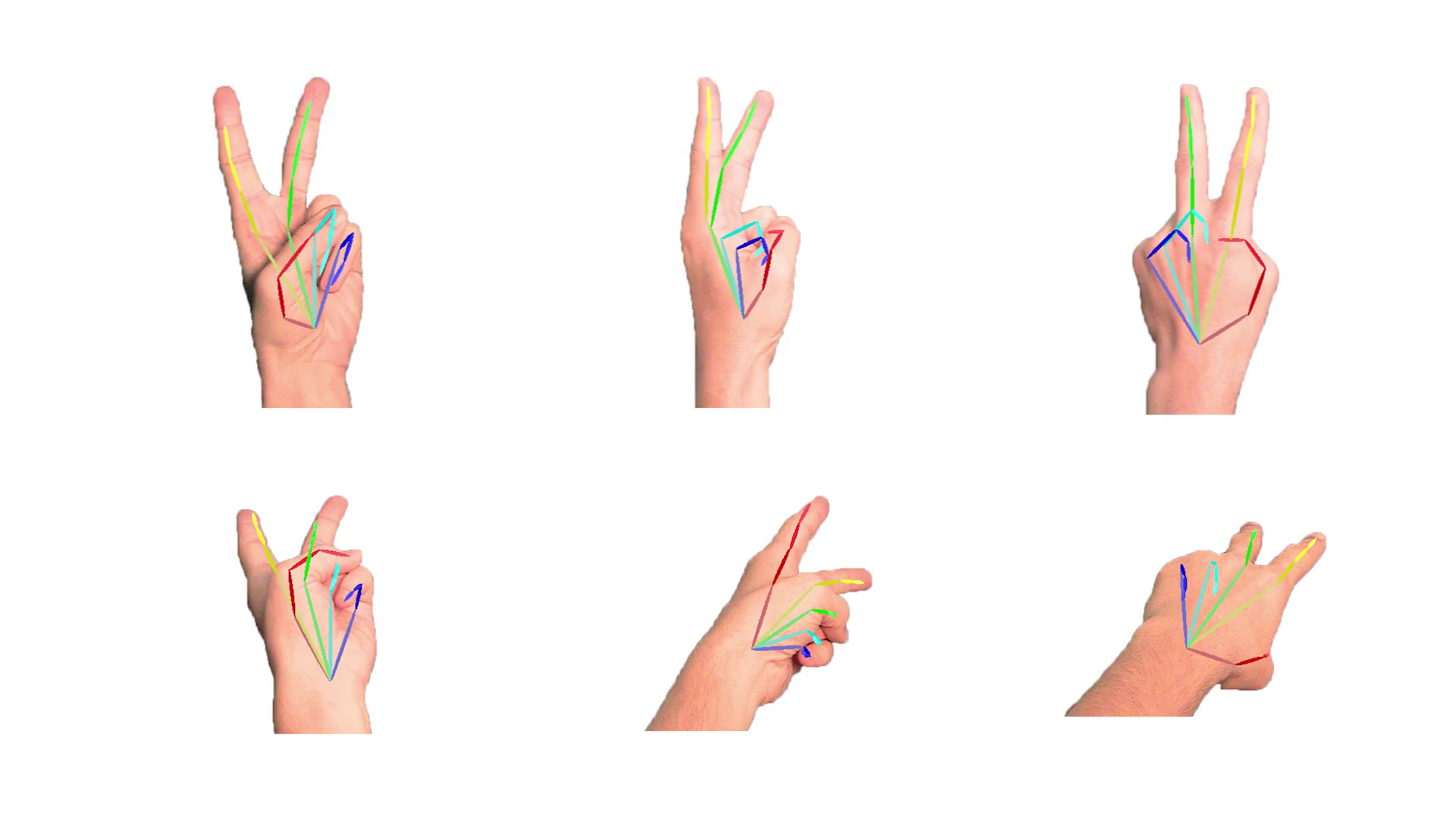}
\caption{Pictures of six hands all performing the same hand shape (v-shape) taken from six different orientations. Mediapipe fails at estimating the pose of the bottom-middle image.}
\label{fig:original_hands_poses}
\end{figure}

\item \textbf{3D Rotation} The pose is then rotated in 3D space such that the normal of the back of the hand aligns with the $Z$-axis. As a result, the palm plane now resides within the $XY$ plane (Figure \ref{fig:3d_rotation_normalization}).

\begin{figure}[h!]
\centering
\includegraphics[width=\textwidth]{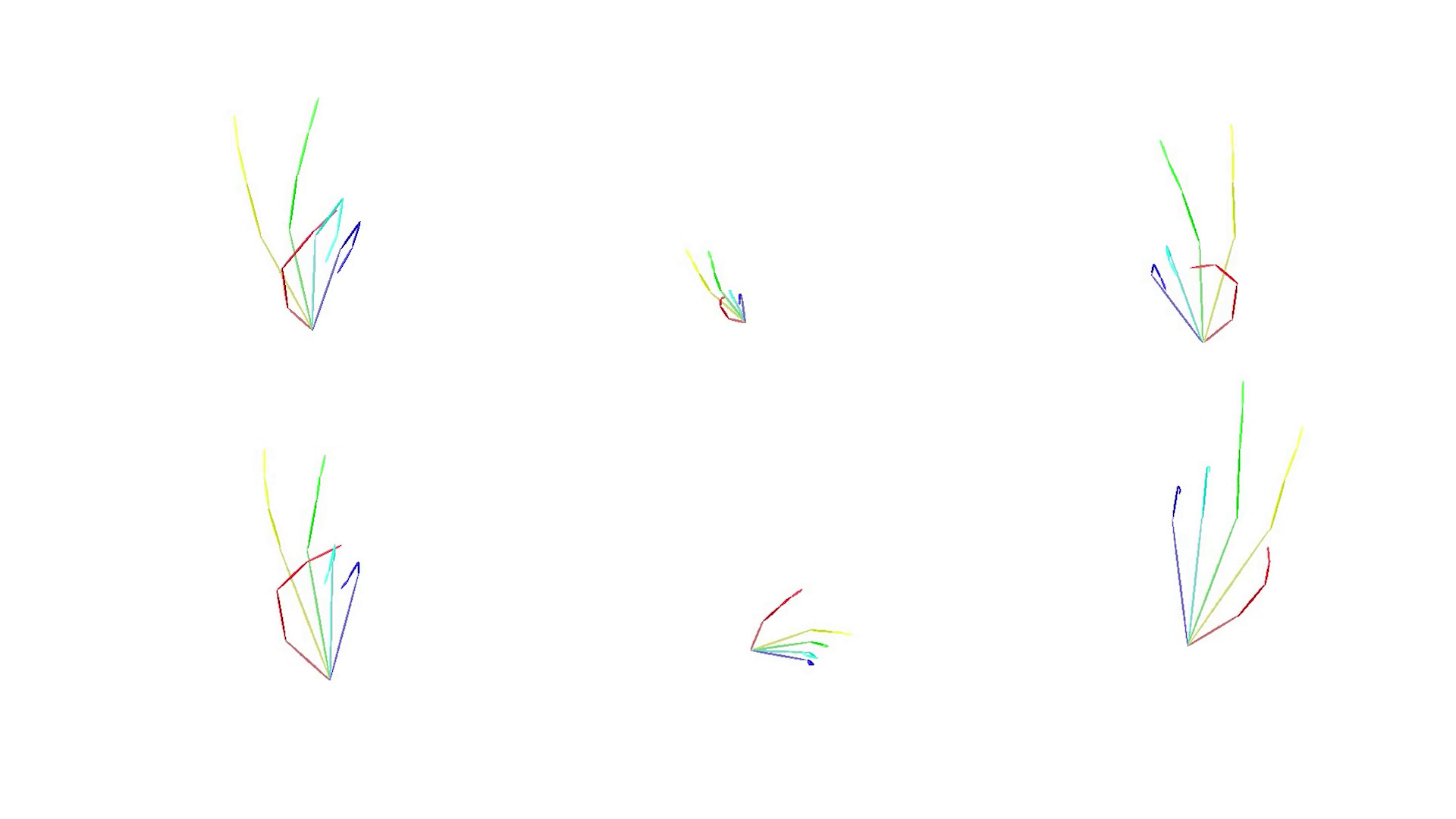}
\caption{Hand poses after 3D rotation. The scale difference between the hands demonstrates a limitation of the 3D pose estimation system used.}
\label{fig:3d_rotation_normalization}
\end{figure}

\item \textbf{2D Orientation} Subsequently, the pose is rotated in 2D such that the metacarpal bone of the middle finger aligns with the $Y$-axis (Figure \ref{fig:2d_rotation_normalization}).

\begin{figure}[h!]
\centering
\includegraphics[width=\textwidth]{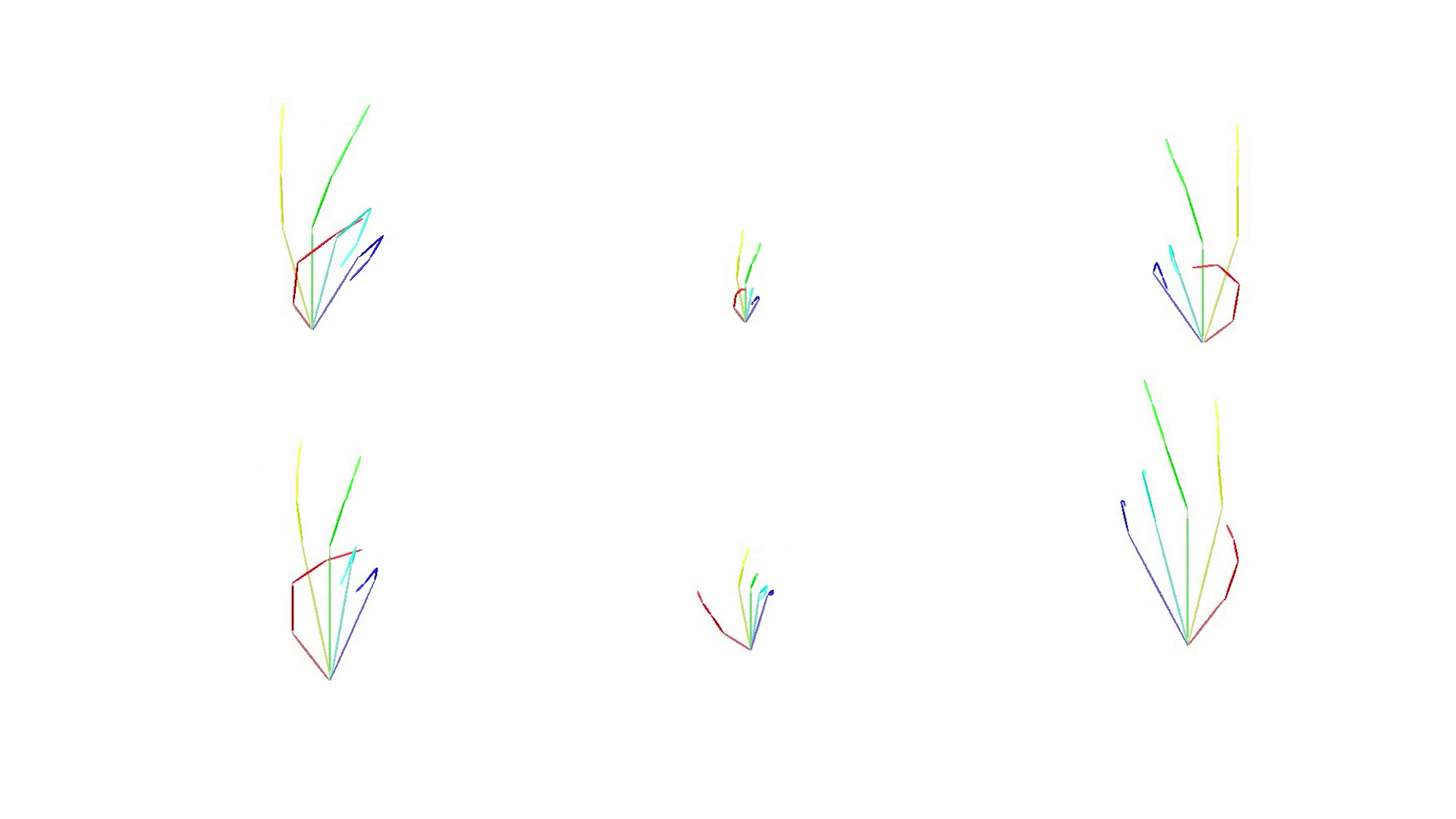}
\caption{Hand poses after being rotated.}
\label{fig:2d_rotation_normalization}
\end{figure}

\item \textbf{Scale} The hand is scaled such that the metacarpal bone of the middle finger attains a constant length (which we typically set to $200$, Figure \ref{fig:scale_normalization}).

\begin{figure}[h!]
\centering
\includegraphics[width=\textwidth]{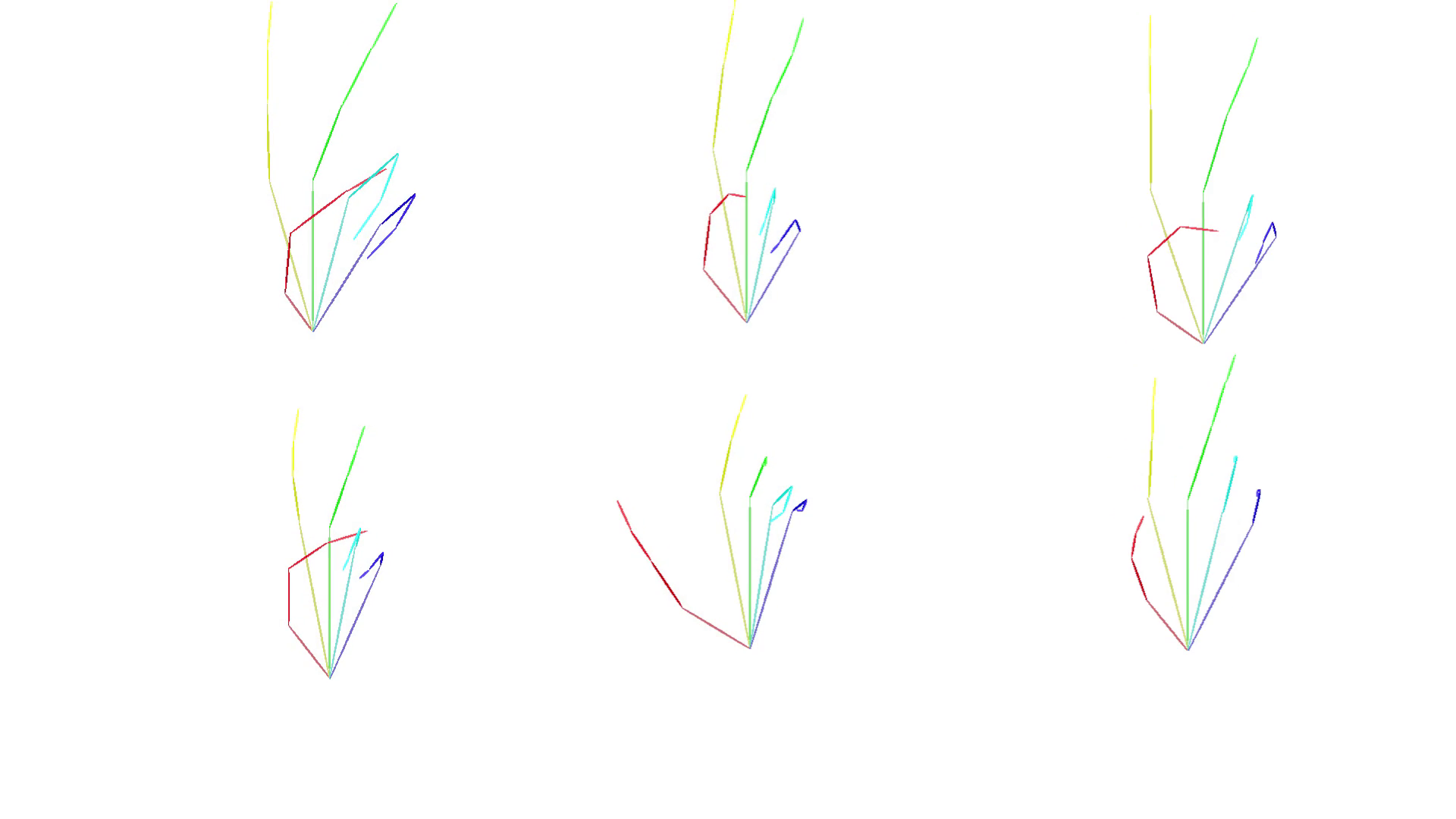}
\caption{Hand poses after being scaled.}
\label{fig:scale_normalization}
\end{figure}

\item \textbf{Translation} Lastly, the wrist joint is translated to the origin of the coordinate system $(0,0,0)$. Figure \ref{fig:translation_normalization} demonstrates how when overlayed, we can see all hands producing the same shape, except for one outlier.

\begin{figure}[h!]
\centering
\includegraphics[width=\textwidth]{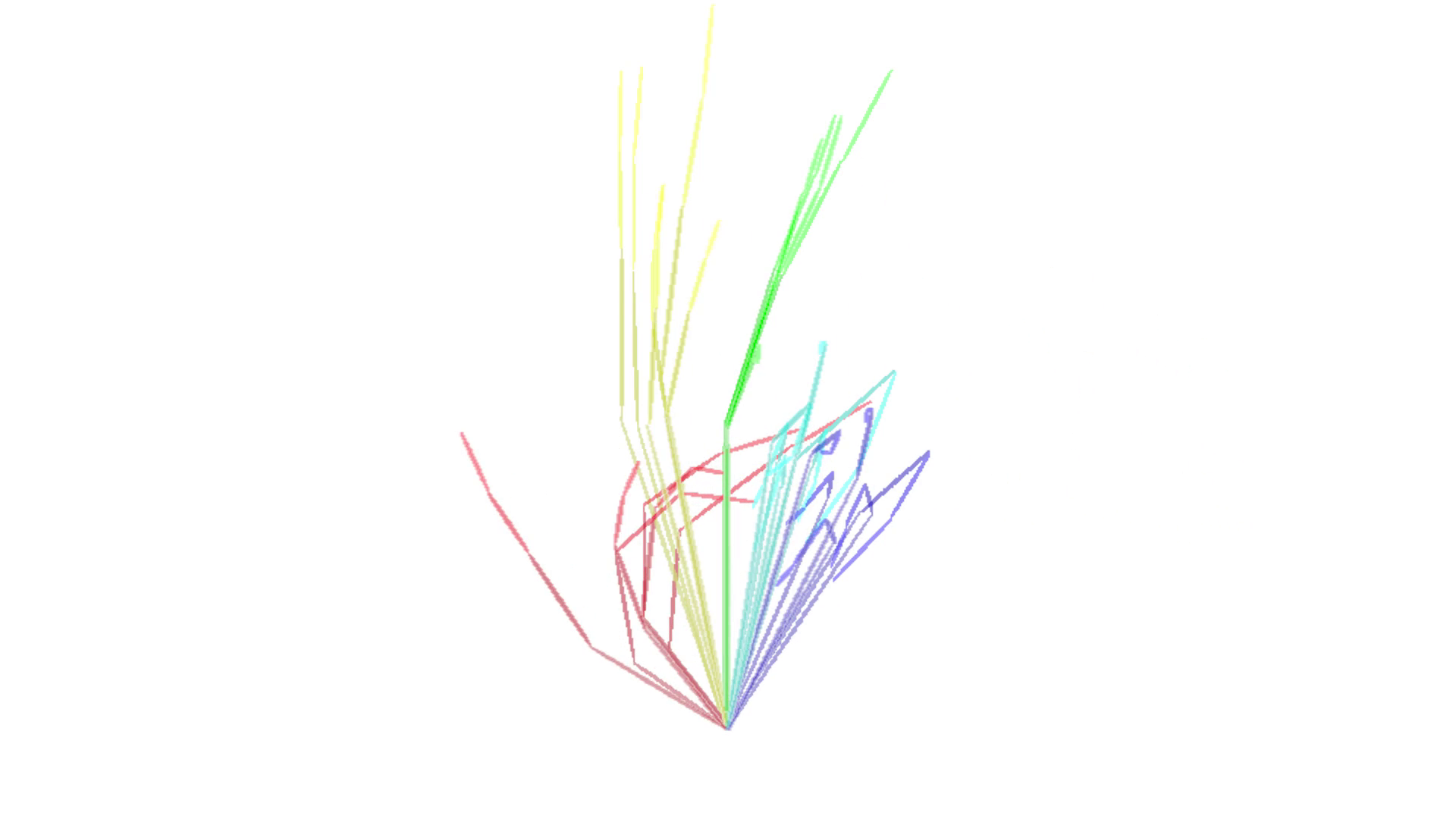}
\caption{Normalized hand poses overlayed after being translated to the same position. The positions of the wrist and the metacarpal bone of the middle finger are fixed.}
\label{fig:translation_normalization}
\end{figure}
\end{enumerate}

By conducting these normalization steps, a hand model can be standardized, reducing the complexity of subsequent steps such as feature extraction and hand shape classification. This standardization simplifies the recognition process and can contribute to improving the overall accuracy of the system.

\subsection{3D Hand Pose Evaluation}

In order to assess the performance of our 3D hand pose estimation and normalization, we introduce two metrics that gauge the consistency of the pose estimation across orientations and crops.

Our dataset is extracted from the SignWriting Hand Symbols Manual \cite{sw-hand-symbols},
and includes images of 261 different hand shapes, from 6 different angles.
All images are of the same hand, of an adult white man.

\paragraph{Multi Angle Consistency Error (MACE)} evaluates the consistency of the pose estimation system across the different orientations. We perform 3D hand normalization, and overlay the hands. The MACE score is the average standard deviation of all pose landmarks, between all views.
A high MACE score indicates a problem in the pose estimation system's ability to maintain consistency across different orientations. This could adversely affect the model's performance when analyzing hand shapes in sign languages, as signs can significantly vary with hand rotation.

\begin{figure}[h!]
    \centering
    \includegraphics[width=0.09\textwidth]{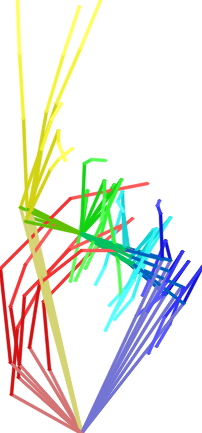} \includegraphics[width=0.09\textwidth]{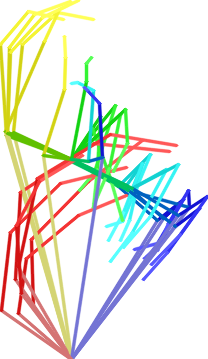} \includegraphics[width=0.09\textwidth]{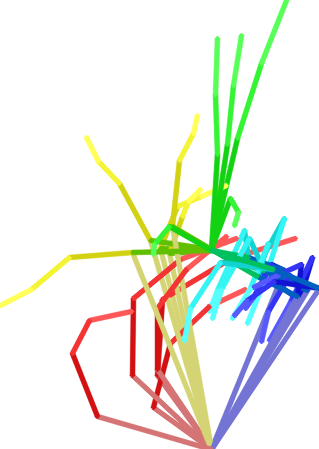} \includegraphics[width=0.09\textwidth]{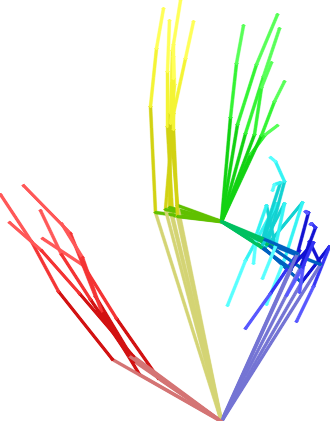} \includegraphics[width=0.09\textwidth]{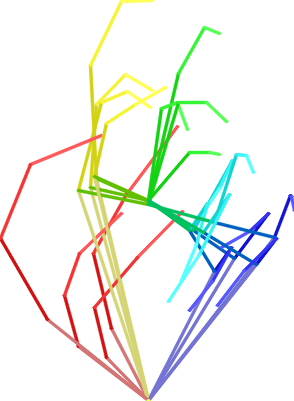}
    \includegraphics[width=0.09\textwidth]{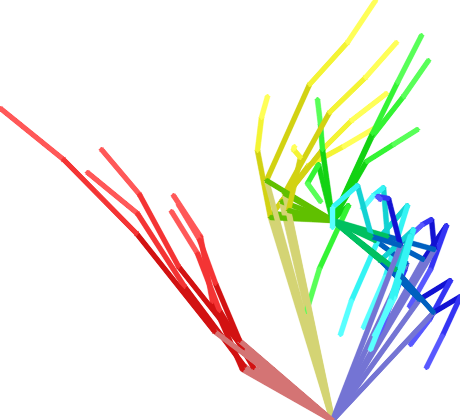} \includegraphics[width=0.09\textwidth]{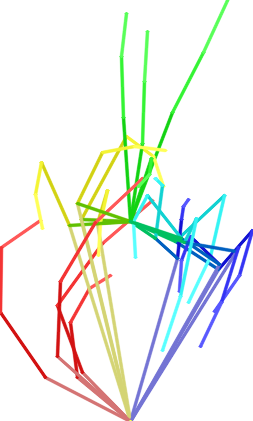} \includegraphics[width=0.09\textwidth]{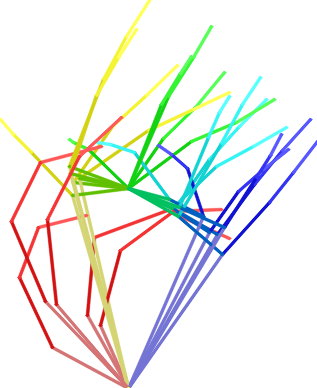} \includegraphics[width=0.09\textwidth]{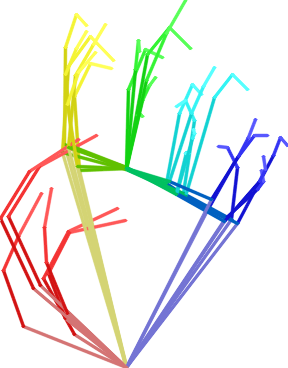} \includegraphics[width=0.09\textwidth]{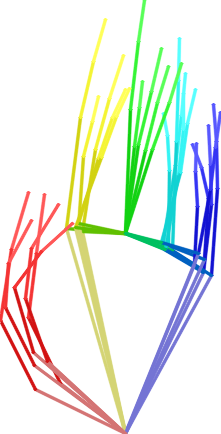}
    \caption{Visualizations of 10 hand shapes, each with 6 orientations 3D normalized and overlayed.}
    \label{fig:mace-example}
\end{figure}

Figure \ref{fig:mace-example} shows that our 3D normalization does work to some extent using Mediapipe. We can identify differences across hand shapes, but still note high variance within each hand shape. 

\paragraph{Crop Consistency Error (CCE)} gauges the pose estimation system's consistency across different crop sizes. We do not perform 3D normalization, but still overlay all the estimated hands, shifting the wrist point of each estimated hand to the origin $(0,0,0)$. The CCE score is the calculated average standard deviation of all pose landmarks across crops.
A high CCE score indicates that the pose estimation system is sensitive to the size of the input crop, which is a significant drawback as the system should be invariant to the size of the input image.

\begin{figure}[h!]
    \centering
    \includegraphics[width=0.09\textwidth]{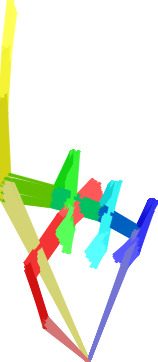} \includegraphics[width=0.09\textwidth]{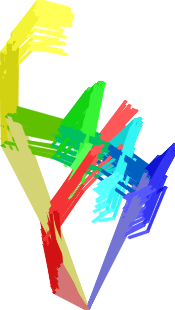} \includegraphics[width=0.09\textwidth]{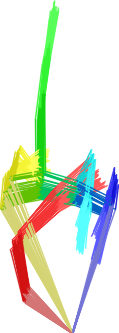} \includegraphics[width=0.09\textwidth]{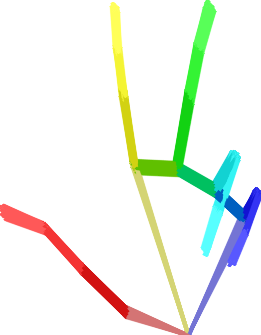} \includegraphics[width=0.09\textwidth]{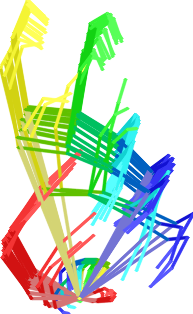}
    \includegraphics[width=0.09\textwidth]{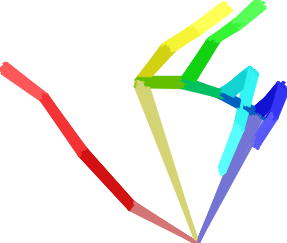} \includegraphics[width=0.09\textwidth]{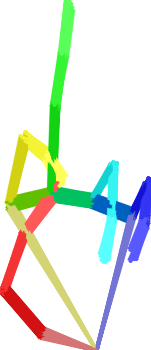} \includegraphics[width=0.09\textwidth]{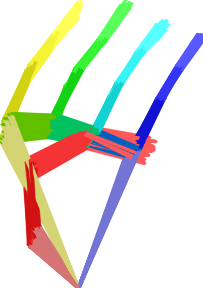} \includegraphics[width=0.09\textwidth]{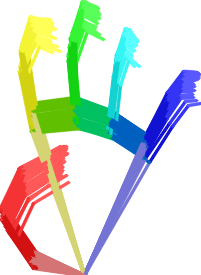} \includegraphics[width=0.09\textwidth]{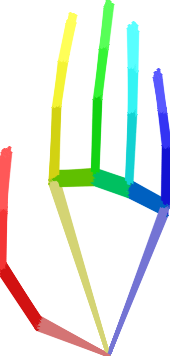}
    \caption{Visualizations of 10 hand shapes, each with 48 crops overlayed.}
    \label{fig:cce-example}
\end{figure}

Figure \ref{fig:cce-example} shows that for some poses, Mediapipe is very resilient to crop size differences (e.g. the first and last hand shapes). However, it is concerning that for some hand shapes, it exhibits very high variance, and possibly even wrong predictions.

\subsection{Conclusion}

Our normalization process appears to work reasonably well when applied to different views within the same crop size. It succeeds in simplifying the hand shape, which in turn, can aid in improving the accuracy of hand shape classification systems.

However, it is crucial to note that while this method may seem to perform well on a static image, its consistency and reliability in a dynamic context, such as a video, may be quite different. In a video, the crop size can change between frames, introducing additional complexity and variance. This dynamic nature coupled with the inherently noisy nature of the estimation process can pose challenges for a model that aims to consistently estimate hand shapes.

In light of these findings, it is clear that there is a need for the developers of 3D pose estimation systems to consider these evaluation methods and strive to make their systems more robust to changes in hand crops. The Multi Angle Consistency Error (MACE) and the Crop Consistency Error (CCE) can be valuable tools in this regard. 

MACE could potentially be incorporated as a loss function for 3D pose estimation, thereby driving the model to maintain consistency across different orientations. Alternatively, MACE could be used as an indicator to identify hand shapes that require more training data. It is apparent from our study that the performance varies greatly across hand shapes and orientations, and this approach could help in prioritizing the allocation of training resources.

Ultimately, the goal of improving 3D hand pose estimation is to enhance the ability to encode signed languages accurately. The insights gathered from this study can guide future research and development efforts in this direction, paving the way for more robust and reliable sign language technology.

The benchmark, metrics, and visualizations are available at \url{https://github.com/sign-language-processing/3d-hands-benchmark/}.

\end{document}